%% file: main.tex
\documentclass[10pt,twocolumn,letterpaper]{article}

\usepackage{cvpr}              %

\usepackage{graphicx}
\usepackage{amsmath}
\usepackage{amssymb}
\usepackage{booktabs}
\usepackage{graphicx}
\usepackage{url}
\usepackage{algorithm}
\usepackage{algorithmicx}
\usepackage{algpseudocode}
\usepackage{nicefrac}
\usepackage{array}
\usepackage{multirow}
\usepackage{mathtools}
\usepackage{cite}
\usepackage{nicematrix}

\usepackage[pagebackref,breaklinks,colorlinks]{hyperref}

\usepackage[capitalize]{cleveref}
\crefname{section}{Sec.}{Secs.}
\Crefname{section}{Section}{Sections}
\Crefname{table}{Table}{Tables}
\crefname{table}{Tab.}{Tabs.}

\def\cvprPaperID{2185} %
\def\confName{CVPR}
\def\confYear{2023}

\begin{document}

\title{Batch Model Consolidation: A Multi-Task Model Consolidation Framework}

\author{Iordanis Fostiropoulos \hspace{20pt} Jiaye Zhu \hspace{20pt} Laurent Itti\\
University of Southern California, Los Angeles, United States\\
{\tt\small \{fostirop, jiayezhu, itti\}@usc.edu}
}
\maketitle

\begin{abstract}
In Continual Learning (CL), a model is required to learn a stream of tasks sequentially without significant performance degradation on previously learned tasks. Current approaches fail for a long sequence of tasks from diverse domains and difficulties. Many of the existing CL approaches are difficult to apply in practice due to excessive memory cost or training time, or are tightly coupled to a single device. With the intuition derived from the widely applied mini-batch training, we propose Batch Model Consolidation (\textbf{BMC}) to support more realistic CL under conditions where multiple agents are exposed to a range of tasks. During a \textit{regularization} phase, BMC trains multiple \textit{expert models} in parallel on a set of disjoint tasks. Each expert maintains weight similarity to a \textit{base model} through a \textit{stability loss}, and constructs a \textit{buffer} from a fraction of the task's data. During the \textit{consolidation} phase, we combine the learned knowledge on `batches' of \textit{expert models} using a \textit{batched consolidation loss} in \textit{memory} data that aggregates all buffers. We thoroughly evaluate each component of our method in an ablation study and demonstrate the effectiveness on standardized benchmark datasets Split-CIFAR-100, Tiny-ImageNet, and the Stream dataset composed of 71 image classification tasks from diverse domains and difficulties. Our method outperforms the next best CL approach by 70\% and is the only approach that can maintain performance at the end of 71 tasks; Our benchmark can be accessed at \url{https://github.com/fostiropoulos/stream_benchmark}
\end{abstract}

\section{Introduction}

\begin{figure}
  \centering
  \includegraphics[width=1.0\linewidth]{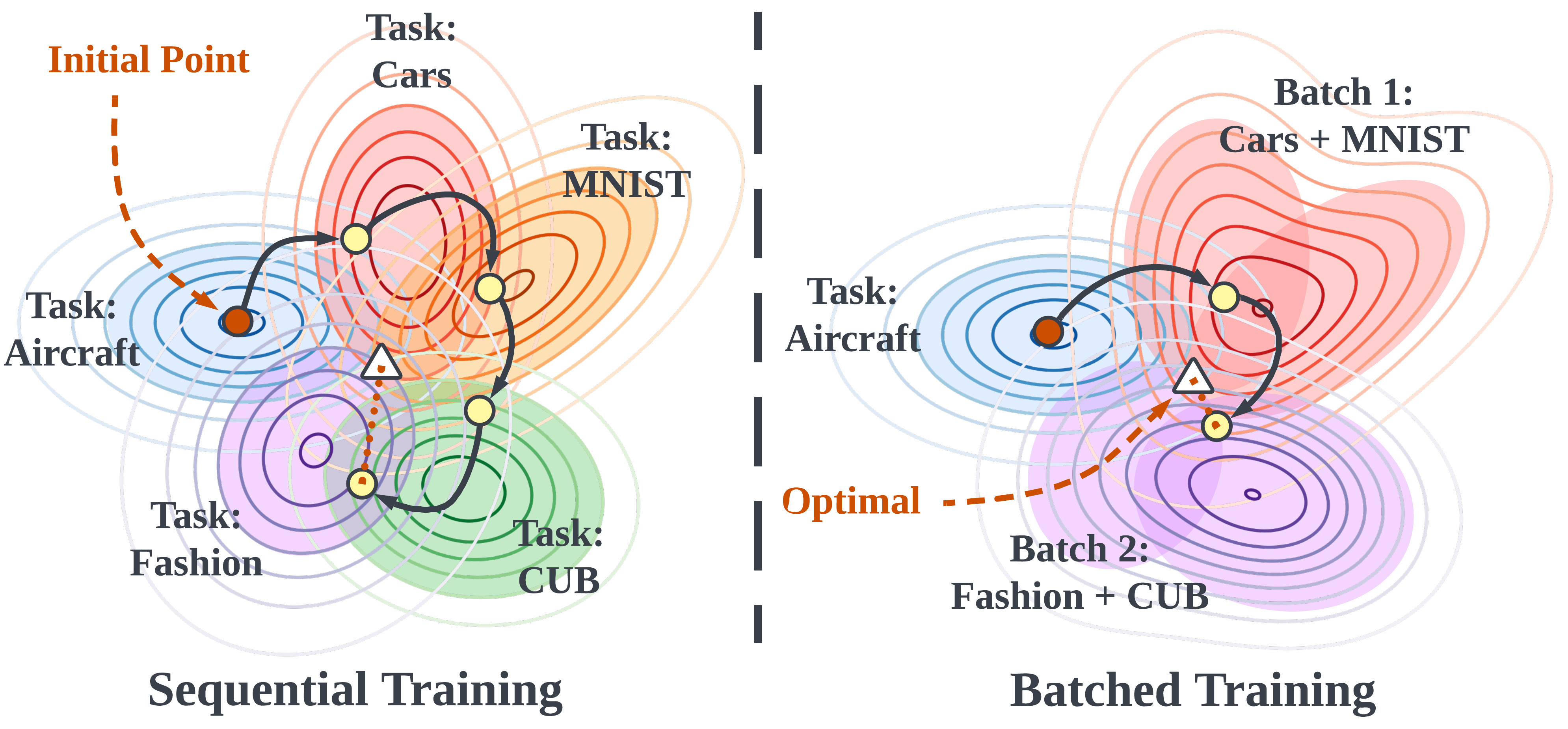}
  \caption{The loss contours by sequential training compared with batch task training \cite{Survey:EmbracingChangeCL} (shaded areas as low-error zones for each task). Intuition: Similar to mini-batch training batched task training can reduce the local minima and improve the convexity of the loss landscape.}
  \label{fig:intuition}
\end{figure}

Continual Learning (CL) has allowed deep learning models to learn in a real world that is constantly evolving, in which data distributions change, goals are updated, and critically, much of the information that any model will encounter is not immediately available \cite{Survey:ClassIncremental}. Current approaches in CL provide a trade-off to the stability-plasticity dilemma \cite{Survey:dilemma} where improving performance for a novel task leads to catastrophic forgetting. %

Continual Learning benchmarks are composed of a limited number of tasks and with tasks of non-distinct domains, such as Split-CIFAR100 \cite{cifar100} and Split-Tiny-ImageNet \cite{tinyimagenet}. Previous approaches in Continual Learning suffer significant performance degradation when faced with a large number of tasks, or tasks from diverse domains \cite{Survey:EmbracingChangeCL}.
Additionally, the cost of many methods increases with the number of tasks \cite{Survey:DefyingForgetting, PNN} and becomes ultimately unacceptable for certain applications, while other methods \cite{EWC, LFL, IMM, SI} are tightly coupled to training on a single device and therefore cannot benefit from scaling in distributed settings. As such, current approaches are impractical for many real-world applications, where multiple devices are trained on a diverse and disjoint set of tasks with the goal of maintaining a single model.

\begin{figure*}[ht]
  \centering
  \includegraphics[width=0.95\linewidth]{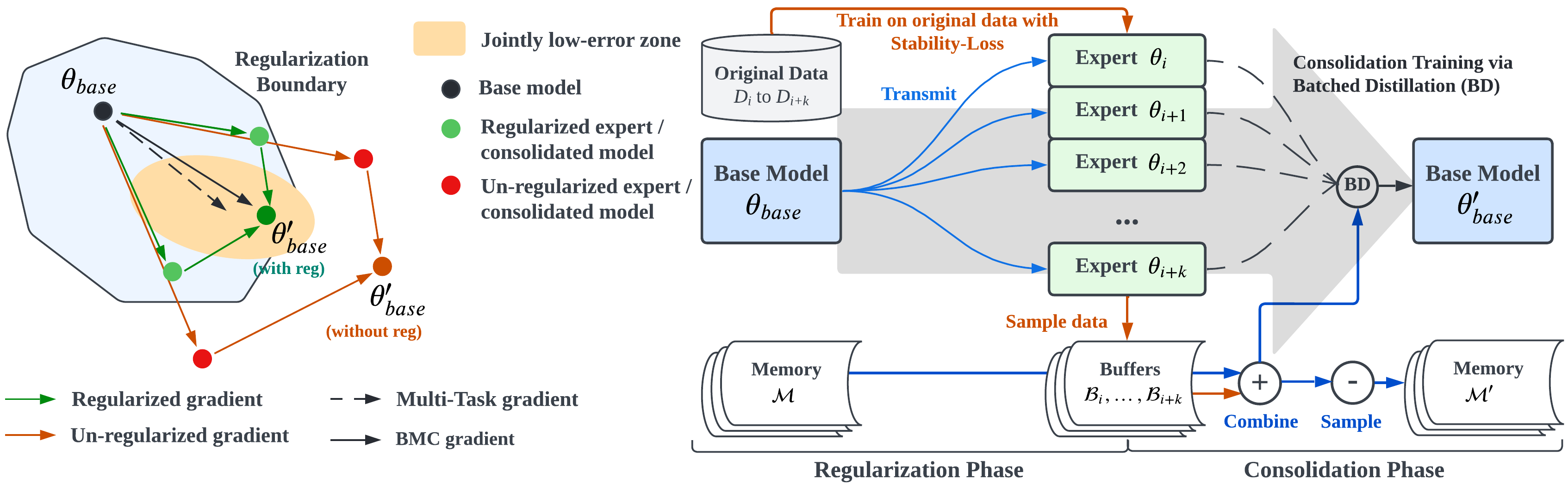}
  \caption{A single incremental step of BMC. On the \textbf{right} figure, the updating of a base model with \textit{Multi-Expert Training}: after receiving the data of the new tasks $D_i, \dots, D_{i+k}$, a batch of experts $\theta_i, \dots, \theta_{i+k}$ are trained separately on their corresponding tasks with \textit{stability loss} applied from the base model. The newly trained experts then sample a subset of their training data and combine them with the memory to perform \textit{batched distillation} on the base model. On the \textbf{left} figure, the regularization helps the batched distillation to update the model closer to the regularization boundary and towards the jointly low-error zone of old tasks and two new tasks.}
  \label{fig:bmc}
\end{figure*}

Motivated by the performance, memory cost, training time and flexibility issues of current approaches,
we propose \textbf{Batch Model Consolidation (BMC)}, a Continual Learning framework that supports distributed training on multiple streams of diverse tasks, but also improves performance when applied on a single long task stream. Our method trains and consolidates multiple workers that each become an expert in a task that is disjoint from all other tasks. In contrast, for Federated Learning the training set is composed of a single task of heterogeneous data \cite{Survey:Federated}.

Our method is composed of two phases. First, during the \textbf{regularization phase}, a set of \textit{expert models} is trained in new tasks in parallel with their weights regularized to a \textit{base model}. Second, during the \textbf{consolidation phase} the expert models are combined into the base model in a way that better retains the performance on the current tasks of all experts and all previously learned tasks. %
The main advantage of our method is that it provides a better approximation to the \textit{multi-task gradient} of all tasks from all expert models, \cref{fig:bmc}. Lastly, BMC better retains performance for significantly more tasks than current baselines, while reducing the total time of training when compared to training on the same task-stream in a sequential manner. The primary contributions of our paper are as follows.
\begin{enumerate}
\item We propose Batch Model Consolidation (\textbf{BMC}) to support CL for training multiple \textit{expert models} on a single task stream composed of tasks from diverse domains.
\item We extend \textit{BMC} for a distributed learning framework where we train multiple \textit{expert models} on disjoint task streams.
\item We propose a \textit{stability loss} to reduce forgetting that is applied between \textit{expert models} and a \textit{base model}. Lastly, a \textit{batched distillation loss} combines multiple \textit{expert models} to update a single \textit{base model} in a single incremental step.
\item We verify our approach on popular benchmarks and show that BMC is robust against large domain-shifts and for a large number of tasks.
\end{enumerate}

\section{Related Works}
\label{sec:related}

Following the taxonomy by De Lange \etal \cite{de2021continual} we summarize methods that mitigate forgetting in three categories, Replay, Regularization and Parameter isolation methods.

\textbf{Replay methods} identify a limited number of exemplars to store in an auxiliary dataset, \textit{buffer}, that is used to retain performance on previously seen tasks through \textit{rehearsal} (ER \cite{ER}, GEM \cite{GEM}, A-GEM \cite{AGEM}, GSS \cite{GSS}). An auxiliary loss can be applied as a regularization term to the main training task, such as with \textit{Knowledge Distillation} (DER++ \cite{DER}, iCaRL \cite{iCaRL}, FDR \cite{FDR}, DMC \cite{DMC}, ExModel \cite{ExModel}) or by restricting the gradient magnitude (GEM \cite{GEM}, A-GEM \cite{AGEM}). Exemplars can be randomly selected from the original dataset \cite{DER} or synthetically generated \cite{ExModel}. Similarly, we perform distillation on previously stored exemplars in a \textit{memory} bank to consolidate knowledge from previous tasks. In contrast to previous works we perform a two-step process of a \textit{regularization-phase} where we maintain proximity of the newly trained task-specific (expert) model to the old (base) model by a \textit{stability loss} as opposed to `knowledge transfer', and in the second \textit{consolidation-phase}, we apply \textit{batched distillation loss} on the pair-wise intermediate representations between \textbf{multiple} expert models and an older model on real exemplars from a buffer. We identify that combining multiple `teacher' models in a single step is better than a single `teacher' model in multiple steps (DMC \cite{DMC}), and we provide a theoretical justification of the result in \cref{sec:gradient_noise}.

Other methods store exemplars in their buffer using prototypes \cite{iCaRL}, increasing exemplar representability \cite{GSS}, or gradient projection \cite{GEM, AGEM} to restrict drastic gradient updates. Such methods are orthogonal to our method, since BMC is extendable to different types of buffer sampling methods and regularization loss as the \textit{stability loss}.

\begin{figure*}
  \centering
  \includegraphics[width=0.87\linewidth]{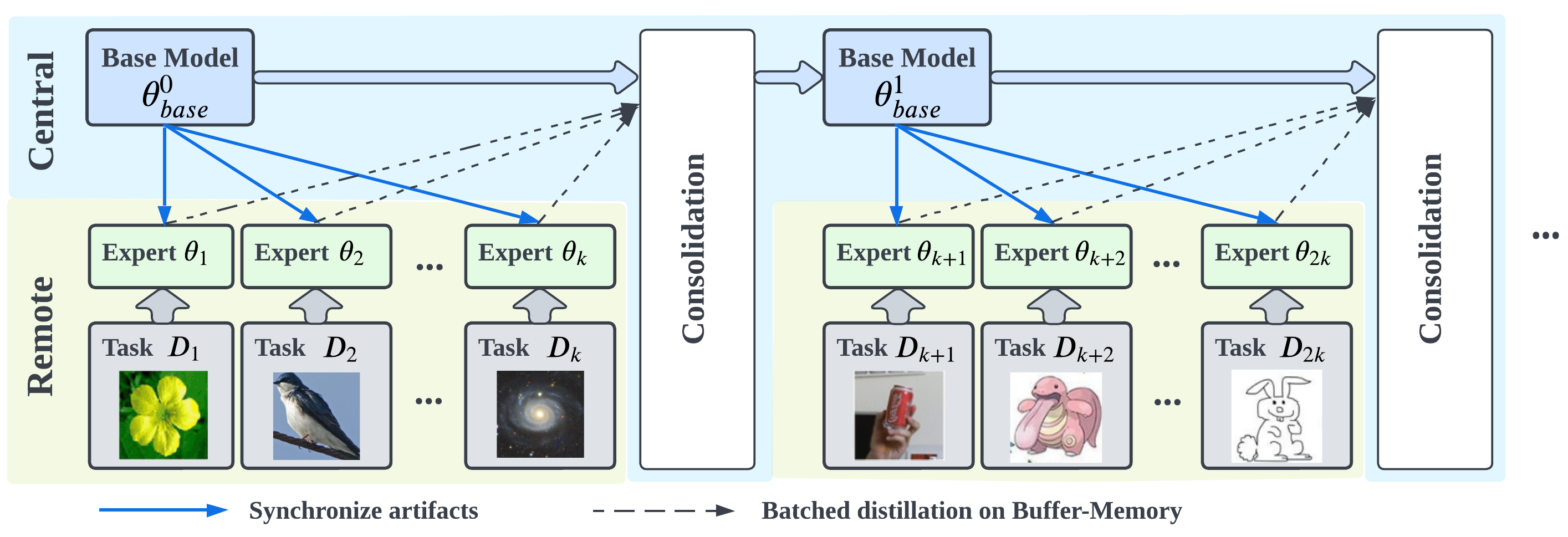}
\caption{Distributed training paradigm of BMC. The central device maintains only a base model and Memory. All experts can be trained in parallel on non-overlapping tasks using distributed devices. Batched Distillation is applied agnostic to the expert model weights on Buffer and Memory data.}
\label{fig:parallel}
\end{figure*}

\textbf{Parameter-isolation} approaches keep the important weights fixed to reduce forgetting. SupSup \cite{SupSup}, HAT \cite{HAT}, PSP \cite{PSP}, PNN \cite{PNN}, and BLIP \cite{BLIP} identify and assign task-specific parameters in the model via \textit{supermasks} or by appending new weights to the model \cite{PNN}. Model Zoo \cite{modelzoo} infers and trains a group of similar tasks into one model as a `weak classifier' to utilize shared domain knowledge, and use an ensemble of models during inference. Such approaches have the number of parameters grow with respect to the number of tasks. Methods such as PackNet \cite{Packnet} and RMN \cite{RMN}, overwrite unimportant parameters to provide larger model capacity for new tasks and do not grow indefinitely. Similarly, we assign an \textit{expert model} to each task to isolate the parameters. However, we maintain the number of experts at each incremental step fixed so that the cost of our method remains constant as the number of tasks grows. Finally, we perform inference using a single \textit{base} model as opposed to an ensemble of models.

\textbf{Regularization methods} such as EWC \cite{EWC} and similarly (MAS \cite{MAS}, SI \cite{SI}) use an auxiliary loss term to constrain optimization \wrt to a metric of importance for each parameter for a given task. LwF \cite{LWF} distills knowledge from the previous model using current task data, and LFL \cite{LFL} freezes portion of the network while penalizing intermediate representations using the Euclidean distance. These approaches are orthogonal to our method and are candidates for the \textit{stability loss}. We find that they underperform compared to our method in \cref{sect:ablation} and additional experiments in the supplementary.

\textbf{Distributed Continual Learning} combines Continual Learning and Federated Learning by incrementally learning a model stored on a central server using distributed devices. Previous works learn the same task on multiple distributed devices that are allocated different subsets of data (CFeD \cite{CFeD}); or train the same sequence of tasks on each device with inter-client communication of model parameters (FedWeIT \cite{FedWeIT}). Our method also combines the knowledge of multiple models trained on distributed devices. In contrast to previous work, we train a unique task on each device, to learn an \textit{expert model}. We consolidate multiple expert models with \textit{batched distillation} on real data as opposed to Knowledge Distillation on an auxiliary dataset \cite{CFeD} or simple aggregation \cite{FedWeIT}. Lastly, in contrast to Federated Learning setting, our method addresses performance constraints as opposed to privacy constraints and prohibits inter-client communication. The remote devices communicate once per incremental step with the central server.

In summary, our approach is a combination of a Regularization method with the use of the stability loss, Replay method with the use of batched distillation loss, and finally Parameter-isolation where we train multiple experts on disjoint tasks and in a distributed fashion.
\section{Preliminaries}
\label{sec:background}
\textbf{Continual Learning} aims to learn a new task from a stream of sequential tasks and without access to previous task datasets, while maintaining performance on all previous tasks. Given a set of Tasks $T = \{T_1 \dots T_n\}$, with clear task boundaries. We train and evaluate our method in a \textbf{Class Incremental Learning (CIL)}\cite{Survey:DefyingForgetting, Survey:ClassIncremental} setting in which task identity is provided during training but not at test time. The Continual Learning objective function for a model with parameters $\theta$, can be summarized as maximizing the average classification accuracy after learning a sequence of $T$ tasks:

\begin{equation} \label{eq:mean_acc}
    \overline{A} = \max_\theta \frac{1}{T} \sum_{t = 1}^T Acc(\theta (T_i), T_{i_y})
\end{equation}

\textbf{Knowledge Distillation (KD)} \cite{KnowledgeDistillation}, in the CL setting, can be used to transfer knowledge between models trained on different tasks \cite{DER,iCaRL}. KD penalizes the student model using a loss function between the representations of teacher and student models. Representations used in KD can be the output logits $\mathcal{L}_{kd}$ \cite{ExModel, DER, DMC, iCaRL} or the intermediate feature vectors $\mathcal{L}_{bd}$ \cite{KD-attention, KD-feature}. Given the hidden representation vectors at depth $i \in |\theta|$ from student and teacher models, $\phi_i^s, \phi_i^t \in \mathbb{R}^d$, we compute $\mathcal{L}_{bd}$ as the sum of the distance between all pair-wise hidden representations. 
\begin{equation} \label{eq:kd_mse}
    \mathcal{L}_{bd} (\theta_t(x), \theta_s(x)) = \sum_{i=1}^{|\theta|} ||sg(\phi_i^t)- \phi_i^s||_2
\end{equation}
$sg$ is the stop gradient operator that prevents the parameters of the teacher model from being updated. %

\textbf{Gradient Noise Reduction}
\label{sec:gradient_noise}
In many nonconvex optimization problems, the loss manifold is filled with local minima and saddle points, where Stochastic Gradient Descent (SGD) optimization can underperform \cite{SGD-minima}. Noise in the training data leads to noise in the gradients and high variance for SGD \cite{GradientNoise}. Gradient approximation methods, such as minibatch training, accumulate gradients over a batch of data to estimate the \textit{true gradient} of the entire training set. Keskar \etal \cite{LargeBatchGenGap} observed that as the batch becomes smaller, the parameters are updated further away from their initial point as opposed to large batch training. This observation is in agreement with \cite{GradientNoise, TrainLonger} that small batch introduces more `randomness' as it is a lesser approximation to the true gradient of the entire training set and causes instability in training.

Similarly, consider a continuum learning environment where there is a set of \textit{expert} models trained on a disjoint set of tasks with the goal of consolidating them sequentially into a single \textit{base} model. We argue that the consolidation training process is similar to batch training but in the multi-task setting, where we reduce the variance by consolidating the experiences from multiple experts. Previous model consolidation methods \cite{ExModel, DMC} combine a single expert at a time. In contrast, for our method, we observe that batched consolidation improves accuracy as well as enables data parallelism that can speed up training.

\section{Method}
\label{sect:method}

\textbf{Batch Model Consolidation (BMC)}, combines a rehearsal-based learning system and a two-step training process, \cref{fig:bmc} (right).
We first introduce the main design components of BMC for a single task sequence, a \textit{task stream}. Next, we formalize the constraints under which we evaluate \textit{Multi-Expert Training}, where multiple experts are trained in parallel on distinct \textit{task streams} composed of sequences with distinct tasks. In short, our method is composed of multiple \textit{training incremental steps} until all tasks are learned. Each step trains multiple expert models in parallel. Each expert is trained on a specific task, different from all other tasks. The training of each expert is composed of a \textbf{regularization phase} that reduces the deviation of the parameters for the current task from the base model. 
At the end of the training of all expert models for the current incremental step, a \textbf{consolidation phase} distills the expert knowledge back to the base model simultaneously using \textbf{batched distillation loss}. Our method performs better as compared to single-model distillation, and is a better approximation to the `multi-task gradient' \cref{fig:bmc} (left) and \cref{fig:intuition}.

\subsection{Buffer-Memory}
\label{sect:memory}

BMC uses a short-term buffer storage and a long-term memory bank to store real samples for rehearsal. \textbf{Memory} is a fixed-size storage that holds training exemplars from multiple previous tasks and is only accessible by the base model. \textbf{Buffer} is a temporary storage of limited size for a subset of the expert's training data. For each \textit{train incremental step} of $k$ experts, at the end of the \textit{regularization phase}, the central memory bank is combined with $\mathcal{B}_1$, ..., $\mathcal{B}_k$ from experts. At the end of the \textit{consolidation phase}, the memory is subsampled to maintain a constant size. 

\textbf{Sampling methods} are applied for both memory and buffer data selection to meet the size constraints of each storage solution. The goal of a sampling method is to improve the informativeness of the buffer data for the current task and the memory data for all previously learned tasks. We experiment with multiple sampling methods, including gradient-based sampling \cite{GSS} and random selection in an ablation study \cref{sect:ablation}. 

\subsection{Stability Loss} 
\label{sec:regularization}

During the Regularization phase, we train an \textbf{expert} model $\theta_{exp}$ that is initialized from a \textbf{base} model $\theta_{base}$. Stability loss is applied during the regularization phase and poses a constraint optimization problem during the training of the expert on the new task. The goal of the expert is to learn the new task while maintaining feature similarity to previously learned tasks represented by $\theta_{base}$, implicitly and without access to previous task data. The idea follows a direct comparison to previous Continual Learning regularization-based approaches \cite{oEWC, MAS, LWF, SI}. The intuition of the stability loss is to make the model less prone to \textit{task-recency bias}  \cite{Survey:ClassIncremental} which can be viewed as the root cause of forgetting. Additionally, our ablation studies support the view that the stability loss improves consolidation as each expert's weights are confined within the regularization boundary (\cref{fig:bmc}). The optimization objective of each expert can be summarized as: 
\begin{equation} \label{eq:reg_loss}
    \mathcal{L}_{exp} = \mathcal{L}_{T} (\theta_{exp} (x), y) + \lambda \mathcal{L}_{bd} (\theta_{base} (x),\theta_{exp} (x))
\end{equation}
where $ \mathcal{L}_{T}$ is the task loss and $\mathcal{L}_{bd}$ is the distillation loss applied with the base model as the `teacher' and the `expert' model as the student and $\lambda$ is the stability coefficient. EWC, or KD on the logits can be applied in direct replacement to $\mathcal{L}_{bd}$. We find experimentally that they under-perform compared to $\mathcal{L}_{bd}$ and discuss the ablation experiment results in \cref{sec:experiments} with additional experiments in the supplementary.

\subsection{Consolidation Phase} 
\label{sec:consolidation}

At the end of training and for all $k$ experts, BMC consolidates the learned knowledge of all experts in a single training step using a batched distillation loss. Batched distillation loss is applied only to the most recent task buffer data $\mathcal{B}$ and to memory $\mathcal{M}$ such that $\mathcal{D} = \{\mathcal{M}, \mathcal{B}_1, \dots, \mathcal{B}_k\}$. 

Instead of performing Knowledge Distillation on a single model or task at a time, batched distillation loss is applied with randomly sampled buffer data and expert representations from $\mathcal{D}$. As such, each training batch for the base model can contain randomly sampled tasks from multiple domains. We hypothesize that batched distillation loss, $\mathcal{L}_{bmc}$, improves performance by improving the convexity of the loss landscape for all tasks, \cref{fig:intuition}. $\mathcal{L}_{bmc}$ penalizes on the difference in feature representations from $\theta'_{base}$ to all experts $\mathcal{E} = \{\theta_1, \dots, \theta_k\}$. 
\begin{equation} \label{eq:bd_loss}
    \mathcal{L}_{bmc} = \sum_{\mathclap{\theta_i \in \mathcal{E}}} \mathbb{E}_{x, \phi (x ; \theta_i) \sim \mathcal{D}} [\mathcal{L}_{bd} (\theta'_{base}(x),  \theta_i(x))]
\end{equation}

We experiment with alternatives to $\mathcal{L}_{bd}$ when computing $\mathcal{L}_{bmc}$. We find that $\mathcal{L}_{bd}$ performs the best with experimental results in \cref{sec:experiments} and the supplementary. The final optimization objective of the base model is the joint optimization of $\mathcal{L}_{bmc}$ and the \textit{experience replay} task loss. The training objective of the base model can be summarized as:
\begin{equation} \label{eq:base_loss}
    \mathcal{L}_{base} = \alpha \mathcal{L}_{T} (\theta'_{base} (x), y) + \beta \mathcal{L}_{bmc} (\theta'_{base}, \mathcal{D})
\end{equation}
where $(x, y) \in \mathcal{D}$, $\alpha$ is the experience replay task loss coefficient and $\beta$ the consolidation coefficient. %

\subsection{Multi-Expert Training}
\label{sec:method_distributed}

We formalize the storage and communication constraints under which we evaluate our method. Multi-Expert Training involves multiple devices trained on disjoint tasks with the goal of learning and maintaining a single model that can perform well on all tasks. Thus, a method is evaluated on the cost under which it can mitigate forgetting and penalizes methods that can grow indefinitely with the number of tasks.

Each train incremental step begins with a \textit{synchronization} phase where the base model initializes each expert with the weights of the base model $\theta_{base}$. Next, and during the \textit{consolidation} phase, the expert models communicate the consolidation artifacts to the central device. We evaluate and report the performance of the updated base model at the end of the \textit{consolidation phase}. 

What is shared between each Expert and the Base model is a flexible design choice that is penalized by the \textbf{Communication cost} $\mathbb{B}_c$, as the total number of bytes sent between the central and remote devices in each incremental training step and the \textbf{Memory cost} $\mathbb{M}_c$ of data maintained on the central device. We use \textbf{Cost Accuracy} $\mathbb{A}_c$ to evaluate a method, on the marginal rate of substitution of mean accuracy, \ref{eq:mean_acc} to the \textbf{Total Cost} $\mathbb{T}_c$ at the given evaluation point
\begin{equation}
\label{eq:accuracy_cost}
\mathbb{A}_c = \frac{\overline{A}}{\mathbb{T}_c}
\end{equation}
 where $\mathbb{T}_c=\overline{\mathbb{M}_c\mathbb{B}_c} + \theta$. In detail, for a given train incremental step, we pose the following constraints:
\begin{itemize}

    \item Each expert $\theta_i$ represents a remote device that has access only to the current task data $T_i$ and $\theta_{base}$.
    \item Each remote device can communicate once at the end of the training process to the central device consolidation artifacts $\mathbb{B}$.
    \item The central device must use the consolidation artifacts and $\mathcal{M}$ to update $\theta_{base}'$

\end{itemize}

Thus, the constraints use $\mathbb{M}_c$ to penalize a method that can naively isolate parameters for each task or store all train artifacts at the end of a consolidation phase. Furthermore, the constraints use $\mathbb{B}_c$ to penalize methods that can transmit a buffer that is similar in size to the task dataset on which the expert was trained. The communication constraints can provide a Pareto front of a method in how well it utilizes the available data. As a real-world example, consider the scenario where a fleet of autonomous vehicles are trained on geographical regions of diverse and disjoint features \cite{multi-task-disjoint}, such as weather patterns and road conditions. Additionally, communication of training data can be prohibitively expensive, and storage of Buffer data or models infeasible for a large fleet of vehicles. \cref{fig:parallel} provides an illustration of the Multi-Expert training and \cref{alg:algorithm_distibuted} the pseudo-code for our method.

\begin{algorithm}[t]
\caption{Distributed Batch Model Consolidation}
\label{alg:algorithm_distibuted}
\textbf{Central Device}:
\begin{algorithmic}[1] %
\State Initialize base model $\theta_{base}$ and memory $\mathcal{M}$
\For{batch of $k$ tasks $\in T$}
    \State \textbf{Synchronization phase} Transmit  $\theta_{base}$
    \State Run Remote Train
    \State \textbf{Consolidation phase} Transmit  $\mathbb{B}$
    \State Train $\theta'_{base}$ using $\mathbb{B}$ by \cref{eq:base_loss}
    \State $\mathcal{M}' \gets \text{sample}(\mathcal{M} \cup \mathcal{B})$
    \State Discard $\mathbb{B}$
\EndFor
\end{algorithmic}
\textbf{Remote Train}:
\begin{algorithmic}[1] %
\State Given current task dataset $T_i$ and $\theta_{base}$
\State Initialize expert $\theta_i \gets \theta_{base}$
\State Train $\theta_i$ on $T_i$ by \cref{eq:reg_loss}
\State Sample $\mathcal{B}_i$ from $T_i$
\State Transmit $\mathcal{B}_i$
\end{algorithmic}
\end{algorithm}

\section{Experiments}
\label{sec:experiments}
We evaluate our method on three Continual Learning benchmarks, Tiny-ImageNet \cite{tinyimagenet} split into 10 tasks, CIFAR-100 \cite{cifar100} split into 10 tasks and 20 tasks. Next, we evaluate our method on a long sequence of diverse tasks to demonstrate BMC's advantage. We use the Stream dataset composed of 71 image classification tasks for rigorous evaluation on average accuracy, Cost Accuracy, and relative training time. Lastly, we evaluate the efficacy of each design component for our method through ablation experiments on Permuted-MNIST, where we train for 128 tasks and a total of 1280 classes. We provide an overview of the dataset in this section and attach additional details in the supplementary. We open-source and provide the extracted feature vectors from the Stream dataset, the code to run the benchmark on the baselines and the code for our method as a Distributed Continual Learning library\footnote{\href{https://github.com/fostiropoulos/stream_benchmark}{https://github.com/fostiropoulos/stream\_benchmark}}.

\textbf{Stream Dataset}.
\label{sec:stream}
Common benchmarks for evaluating Continual Learning methods are built by splitting classes from datasets such as MNIST, CIFAR-10/100 and Tiny-ImageNet, which have subtasks in similar domains, of similar size and number of classes. We aim to evaluate BMC in a more general setting where there are larger domain-shifts, for significantly more tasks that range in difficulty and problem size. Lastly, synthetically generated datasets such as permuted-MNIST can be poor references to performance in applicable scenarios \cite{Survey:DefyingForgetting}. %
To this end, we use Stream which is composed of 71 publicly available image classification datasets \cite{DS:aircraft, DS:apparel, DS:aptos2019, DS:art, DS:asl, DS:boat, DS:cars, DS:cataract, DS:celeba, DS:colorectal, DS:concrete, DS:core50, DS:cub200, DS:deepweedsx, DS:dermnet, DS:dtd, DS:electronic, DS:emnist, DS:eurosat, DS:event, DS:face, DS:fashion, DS:fer2013, DS:fgvc6, DS:fish, DS:flowers, DS:food101, DS:freiburg, DS:galaxy10, DS:garbage, DS:gtsrb, DS:ham10000, DS:handwritten, DS:histaerial, DS:inaturalist, DS:indoor, DS:intel, DS:ip02, DS:kermany2018, DS:kvasircapsule, DS:landuse, DS:lego, DS:malacca, DS:manga, DS:minerals, DS:office, DS:oriset, DS:oxford, DS:pcam, DS:places365, DS:planets, DS:plantdoc, DS:pneumonia, DS:pokemon, DS:products, DS:resisc45, DS:rice, DS:rock, DS:rooms, DS:rvl, DS:santa, DS:satellite, DS:simpsons, DS:sketch, DS:sports, DS:svhn, DS:textures, DS:vegetables, DS:watermarked, DS:weather, DS:zalando} from the computer vision literature and Kaggle \cite{kaggle}. We concatenate the datasets into a stream of tasks. There are a total of 6,770,722 train images and 743,977 validation images with 2866 classes in Stream, with different numbers of classes for each task. Details on each dataset are attached in the supplementary. To speed up the experiments, we extract feature vectors from a pre-trained CLIP model \cite{CLIP} and used them as input to the model. For both our method and the baselines we use identical train hyper-parameters. We use the hyper-parameters as reported in the original paper for each method, where it is applicable. All experiments use an MLP with Residual connections \cite{resnet} on the extracted CLIP feature vectors. 

\textbf{Baselines}. We follow the methodology and compare with the methods reported in \cite{Survey:DefyingForgetting, DER}. When running a method on the Stream dataset we use implementation by Mammoth \cite{DER} and FACIL \cite{Survey:ClassIncremental}. We report results from each respective paper when they are available or work by \cite{Survey:DefyingForgetting, DER}. We compare our method with ER \cite{ER}, GSS \cite{GSS}, A-GEM \cite{AGEM}, iCaRL \cite{iCaRL}, GDumb \cite{gdumb}, DER++ \cite{DER}, Online EWC \cite{oEWC}, SI \cite{SI}, MAS \cite{MAS} and DMC \cite{DMC} with details of each method on \cref{sec:related}.
We train a naive baseline (\textit{SGD}) without any Continual Learning strategy as a lower-bound. We compute the theoretical upper bound on performance as \textit{Multi-Task} accuracy where we use the mean accuracy of SGD on each task.

\subsection{Results}
\label{sect:result}

\begin{figure*}[ht]
  \centering
  \includegraphics[width=0.9\linewidth]{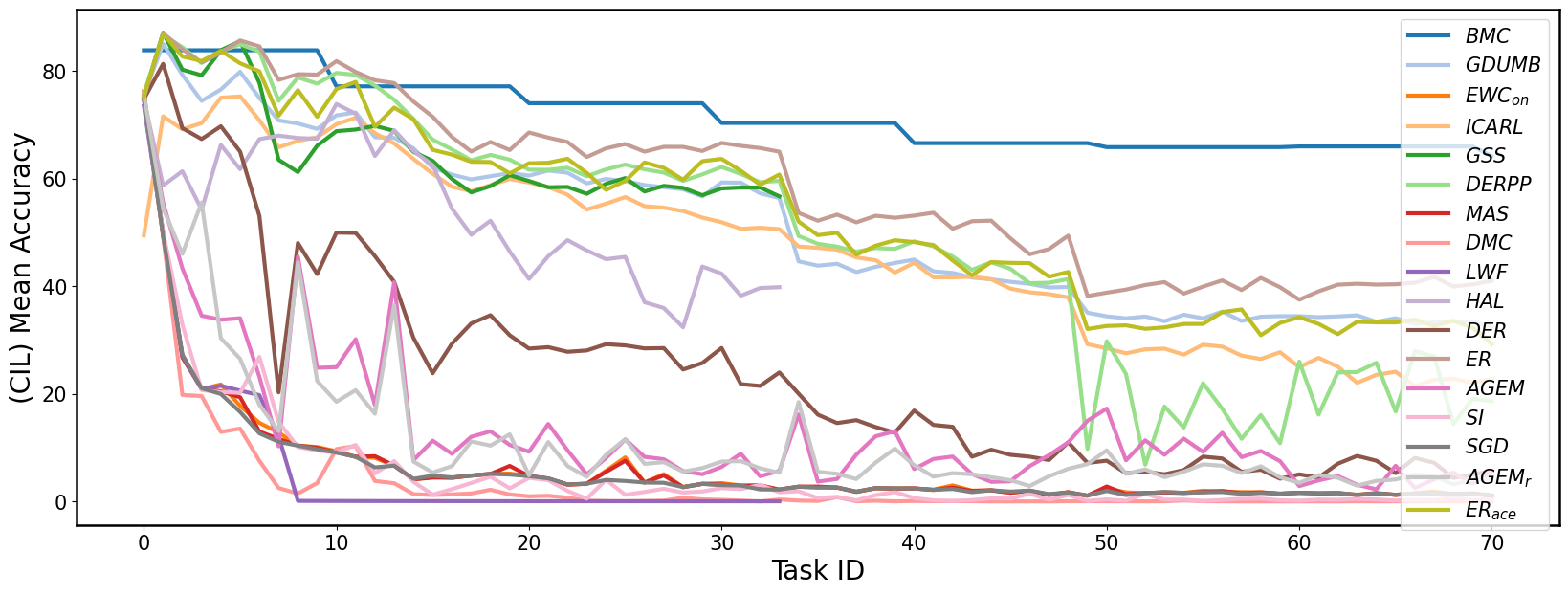}
  \caption{BMC outperforms all other methods on the average accuracy for the Stream dataset and under a CIL evaluation setting.}
  \label{fig:cil_stream}
\end{figure*}

We use 10 experts for our experiments (other training configurations are provided in Supplementary). \Cref{table:benchmark} report the average accuracy $\overline{A}$, for the main baselines. We report Cost Accuracy $\mathbb{A}_c$, Total Cost $\mathbb{T}_c$ and relative training time to SGD. We include additional baselines in \cref{fig:cil_stream} and a full table in the supplementary. For the Continual Learning benchmarks, we show that BMC works well on the short sequence of similar tasks. For the Stream benchmark, our method significantly outperforms all baselines. In detail, BMC outperforms the second highest (ER) by 70\%, and achieve 79\% of the theoretical upper bound provided by Multi-Task training. Additionally, BMC has a constant time complexity \wrt the number of previously seen tasks and is 22\% faster when compared to training with SGD on a single device. Other baselines degrade in relative time performance because they require a second backward propagation \cite{DER} or have an intractable training time as the number of learned tasks increases \cite{DER, GSS, HAL}.

Our experiments and benchmark conclude that most of the recent approaches fail in mitigating forgetting and are outperformed in all regards by simpler alternatives such as ER \cite{ER}. Replay methods that use heuristics in sampling \cite{DER, AGEM, iCaRL} are unable to address the drastic domain-shift of a long stream of tasks and result in sudden performance degradation for large domain-shifts as shown in \cref{fig:cil_stream}. Likewise, Regularization methods \cite{oEWC, MAS} perform similarly to a naive baseline, SGD. Parameter-isolation methods \cite{PNN, SupSup} fail to train due to memory requirements.

We analyze the Total Cost ($\mathbb{T}_c$) as the space complexity of the buffer, memory and any auxiliary information specific to the method. We denote by $|x|$ the input dimension,  $|\theta|$ the number of model parameters, $|\phi|$ the total size of the intermediate representations ($|\phi_{-2}|$ the penultimate feature size, and $|\phi_{-1}|$ the logit size). iCaRL uses a \textit{herding} buffer strategy that computes artifacts for the entire dataset to subsample, which results in higher than theoretical peak memory usage. For our method, we use an efficient implementation to calculate $\mathcal{L}_{bmc}$ that does not transmit $\phi$. As such, BMC maintains a memory footprint per task $O(|x|)+k\theta$.

We compute $\mathbb{T}_c$ in megabytes (MB) for a given buffer size, and thus $\mathbb{A}_c$ represents the improvement in accuracy per unit of MB \cref{eq:accuracy_cost}. Our best performing model variant achieves an accuracy of 71.87\% with a Buffer and Memory size of 15k respectively. For our method we find the Pareto optimal configuration with regards to $\mathbb{A}_c$ and use Memory and Buffer size of 15k and 10k respectively, \cref{table:benchmark}. When comparing between methods, there are limitations in using $\mathbb{A}_c$ for a single configuration. The comparison for the hyper-parameter range that each method was trained on and might be optimal for $\overline{A}$ but not be Pareto optimal in terms of $\mathbb{A}_c$. To this end, we motivate that the evaluation of $\mathbb{A}_c$ is done for multiple configurations. We provide additional details on the limitation of $\mathbb{A}_c$ in \cref{sec:discussion} and provide the Pareto front of our method in the supplementary that can serve as a baseline for future work.

\begin{table*}[ht]
\centering
\begin{tabular}{|c| c| c| c |c|c|c|c|}
 \hline
 \hline
 \bf Methods & S-CIFAR (10) & S-CIFAR (20) & Tiny (10) & Stream (71)  &$\mathbb{A}_c \uparrow$ & $\mathbb{T}_c$ & Time $\downarrow$
 \tabularnewline 
 \hline
 SGD & 8.5 & 3.7 & 7.9 & 2.1 & - & $O(1)$ & 100\%
 \tabularnewline 
 Multi-Task & 75.79 & 75.79 & 68.89 &  89.3 & - & $O(1)$ & 100\%
 \tabularnewline
 \hline
 ER &  12.4\footnotemark[3] & 14.4\footnotemark[2]  & 27.4\footnotemark[1]  & 41.4 &  5.40   & $O(|x|)$ & 184\%
 \tabularnewline
 DER++ &  27.0\footnotemark[2] & - & 39.0\footnotemark[1] & 19.4 & $0.53$ & $O(|x| + |\phi_{-1}|)$ & 205\%
 \tabularnewline
 A-GEM &  6.5\footnotemark[2]  &  3.6\footnotemark[2] & 8.0\footnotemark[1]  & 6.6 & $0.67$ & $O(|x|) + 2|\theta|$ & 231\%
 \tabularnewline
 iCaRL &  25.5\footnotemark[3]  &  19.2\footnotemark[2] & 14.1\footnotemark[1] & 23.4 & 0.59 & $O(|x| + |\phi_{-2:}|)  + |\theta|$ & 141\% %
 \tabularnewline
 GDumb & 36.0\footnotemark[2]   &  22.1\footnotemark[2] & - & 33.0 & $4.29$ & $O(|x|)$ & 129\%
 \tabularnewline
 GSS &  17.4\footnotemark[2]  &  11.3\footnotemark[2] & - & - & - & $O(|x|)$ & 1203\%
 \tabularnewline
 DMC & 36.2 & - & - & 1.0 & $0.02$ & $O(1) + |\theta|$ & 140\%
 \tabularnewline
 \hline
 $\text{EWC}_{\text{on}}$ & 13.1 & 3.7 & 7.6 & 2.1 & $0.99$ & $O(1) + 2|\theta|$ & 172\%
 \tabularnewline
\hline
 \textbf{Ours} & 66.5\footnotemark[4] & 67.4\footnotemark[4] & 49.4\footnotemark[4] & \textbf{70.4} & $\textbf{6.27}$ & $O(|x|) + k \theta$ & \textbf{78\%} 
 \tabularnewline
 \hline
 \hline
\end{tabular}
\caption[Caption for Results Table]{CIL performance on split CIFAR-100 (S-CIFAR) for 10 and 20 tasks, split Tiny-ImageNet for 10 tasks and Stream Dataset for 71 tasks. Baseline results for S-CIFAR and Tiny-ImageNet are from \cite{DER, Survey:result-cifar100-20, result-cifar100-10-1, ER-Tricks, Survey:ClassIncremental, resnet18-cifar100} with the buffer size annotated next to the reported result as 5120\footnotemark[1], 5000\footnotemark[2], 1000\footnotemark[3] or 2000\footnotemark[4]. We use `-' to denote results that are not available in the literature, not applicable or not feasible to obtain. }
\label{table:benchmark}
\end{table*}

\subsection{Ablation Study and Analysis}
\label{sect:ablation}

\begin{figure*}[ht]
  \centering
  \includegraphics[width=1.0\linewidth]{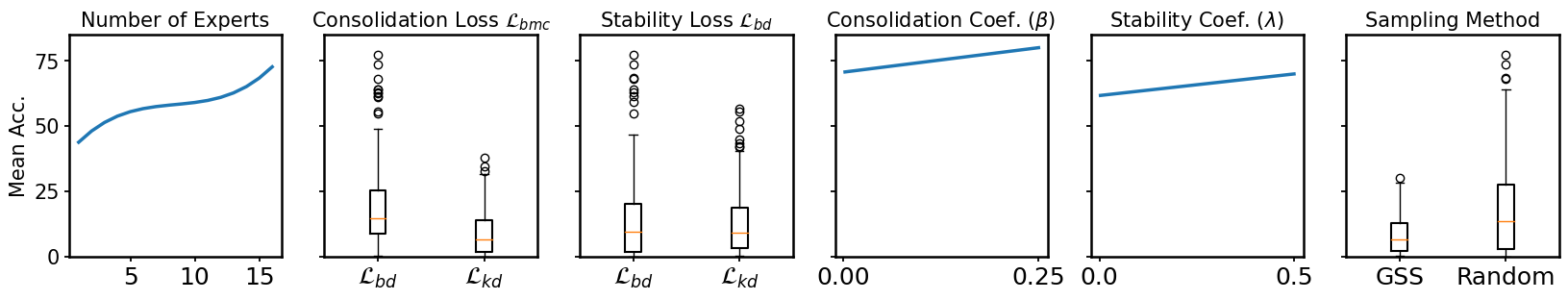}
  \caption{Hyper-parameter ablation results on Permuted-MNIST for 128 tasks and 1280 classes. From left to right, we vary the number of expert models, the loss used in consolidation and \textit{stability loss}, the consolidation and stability coefficients,  and finally the sampling method for constructing the buffer and memory.}
  \label{fig:ablation}
\end{figure*}

We conduct an ablation study on the Permuted-MNIST \cite{mnist} to experimentally verify the impact of each component in BMC. We focus our ablation experiment on different hyper-parameter effects during regularization and consolidation phases and summarize our findings in \cref{fig:ablation}, where the component settings are compared by the mean accuracy. In total we run 629 experiments and for each experiment, we use a different random seed where we uniformly sample values for each hyper-parameter in the reported range. For continuous hyper-parameters and for reasons of clarity, we report the interpolated curve of all random trials. Lastly, for brevity we discuss the most important components in this section and provide experimental results on additional hyper-parameters in the supplementary.

\textbf{Number of experts}. The average accuracy increases monotonically with a larger batch of experts. Consolidating more experts at each step improves convergence on all tasks and reduces gradient noise. Our results agree with our hypothesis in \cref{sec:background} and \ref{sect:method}, where we claim that batched distillation is a better approximation to the joint optimization goal of all tasks.

\textbf{Consolidation \& Stability loss}. We compare the effectiveness of different alternatives for \textit{stability loss} and \textit{batched distillation loss}. We compare our method against Knowledge Distillation (KD) as used by \cite{iCaRL, DER}. In both the regularization and consolidation phases, the pairwise intermediate feature distillation $\mathcal{L}_{bd}$ performed better for BMC. 

\textbf{Stability \& Consolidation coef}. 
Stability coefficient ($\lambda$) scales the stability loss, while the consolidation coefficient ($\beta$) scales $\mathcal{L}_{bmc}$ during the consolidation phase. Results show the final performance is positively correlated to $\lambda$ and $\beta$ with linear correlation coefficients of 4.12e-2 and 9.39e-2 respectively. Our findings agree with our hypothesis that a stronger batched distillation penalty provides a better approximation to the \textit{multi-task gradient} \cref{sec:gradient_noise}.  %

\textbf{Sampling method}. We evaluate more sophisticated sampling strategies, including gradient-based \cite{GSS} and compare with random selection. We found that random sampling significantly outperformed every other sampling method. We argue that more sophisticated methods in sampling are not suitable for a long sequence of tasks as any inductive bias in selecting which samples to store can be biased either to more recent or later tasks and lead to under-represented tasks or classes. Our results are in agreement with the previous survey \cite{Survey:ClassIncremental}.

\textbf{Buffer-Memory size}. We vary the size of buffer and memory to study the effect they have on accuracy. As expected, the performance of our method increases with the increased buffer-memory. We find, however, that the effect of buffer size is greater with a linear corr. coefficient of 0.193 when compared to memory with a coef. of 0.104.

\section{Discussion}
\label{sec:discussion}

Previous work \cite{DER, Survey:result-cifar100-20} shows improvement in standardized benchmarks compared to simpler approaches such as ER \cite{ER}. We reason that the results hold for a short sequence of similar tasks or synthetically generated datasets. In contrast, we find that for the Stream dataset ER outperforms other baseline methods significantly. Works built on top of ER  \cite{DER, iCaRL} that use an auxiliary loss suffer in performance for the same benchmark. The reason for the performance degradation has to be studied further, but we hypothesize that the auxiliary loss introduces the task-recency bias within the artifacts it is applied to. %

Total Cost $\mathbb{T}_c$ and by extension $\mathbb{A}_c$ do not take into consideration of the hyper-parameters used for each method that are flexible design choices and as such can be non-equivalent when comparing two methods and for a single configuration. Consider that the number of parameters for a model can change for a different configuration and can lead to a method with degradation in $\mathbb{A}_c$ if the method stores intermediate gradients \cite{AGEM} or features \cite{iCaRL} compared to methods that only store logits \cite{DER}. For a thorough evaluation between methods that is agnostic to the hyper-parameter a Pareto optimal configuration must be used to evaluate each method. Additionally, an improved version of $\mathbb{A}_c$ that is not influenced by component hyper-parameter choice can provide a better evaluation metric.

\section{Conclusion}

In this paper, we propose Batch Model Consolidation, a Continual Learning framework that reduces catastrophic forgetting when training on a long sequence of tasks from diverse domains and ranging difficulties. Our method combines Regularization, with the use of the stability loss on a previously trained base model; Replay, with the use of batched distillation loss on memory data; and Parameter-Isolation where multiple expert models are trained on a sequence of disjoint tasks. Lastly, we extend our framework to work in a distributed setting where each expert can reside on a different device and specialize in a given task. We experimentally demonstrate that BMC is the only method that maintains performance for our long sequence of 71 tasks. Lastly, we make the code of this work publicly available so that it can serve as a benchmark for future work in Distributed Continual Learning.
\newpage
\section*{Acknowledgement}

This work was supported by C-BRIC (one of six centers in JUMP, a
Semiconductor Research Corporation (SRC) program sponsored by DARPA), DARPA (HR00112190134) and the Army Research Office (W911NF2020053). The authors affirm that the views expressed herein are solely their own, and do not represent the views of the United States government or any agency thereof.

{\small
\bibliographystyle{ieee_fullname}
\bibliography{ref}
}

\include{appendix/supp}

\end{document}

%% file: appendix/supp.tex
\appendix
\section{Experiment Details}

\subsection{Stream Dataset Details}
\label{sec:dataset}
In total, we pre-process and use 71 datasets from the computer vision literature and Kaggle \cite{DS:aircraft, DS:apparel, DS:aptos2019, DS:art, DS:asl, DS:boat, DS:cars, DS:cataract, DS:celeba, DS:colorectal, DS:concrete, DS:core50, DS:cub200, DS:deepweedsx, DS:dermnet, DS:dtd, DS:electronic, DS:emnist, DS:eurosat, DS:event, DS:face, DS:fashion, DS:fer2013, DS:fgvc6, DS:fish, DS:flowers, DS:food101, DS:freiburg, DS:galaxy10, DS:garbage, DS:gtsrb, DS:ham10000, DS:handwritten, DS:histaerial, DS:inaturalist, DS:indoor, DS:intel, DS:ip02, DS:kermany2018, DS:kvasircapsule, DS:landuse, DS:lego, DS:malacca, DS:manga, DS:minerals, DS:office, DS:oriset, DS:oxford, DS:pcam, DS:places365, DS:planets, DS:plantdoc, DS:pneumonia, DS:pokemon, DS:products, DS:resisc45, DS:rice, DS:rock, DS:rooms, DS:rvl, DS:santa, DS:satellite, DS:simpsons, DS:sketch, DS:sports, DS:svhn, DS:textures, DS:vegetables, DS:watermarked, DS:weather, DS:zalando}. We use the train and validation split from each original dataset and for each task. Some datasets have multiple ways to label an image, for example, CelebA \cite{DS:celeba} assigns 40 binary labels to each image, such as `Brown Hair' or `Blurry'. We use `Sub Task' to denote which split or sub-task we use for each dataset and refer the reader to the original dataset documentation on the details of each split. We do not modify or alter any of the datasets in any way. To speed up training and evaluation, we extract the image feature vectors for each dataset using a CLIP model and use them for the classification task directly. \cref{table:dataset} and \cref{table:dataset_2} contain the statistics for each dataset.

\textbf{Task difficulty}. SGD performance can be used as a proxy for task difficulty \cite{Survey:DefyingForgetting}. Some tasks, such as Planets \cite{DS:planets} (task id 50), have really low difficulty and thus high performance on SGD. The inflated performance on the task can be caused by over-fitting or be a reflection of the poor curation process and similarity between the train and validation split of the dataset. The purpose of the Stream dataset is to introduce nuances between tasks, such as with the varying training dataset size. As such, we welcome the nuances and errors that are inherited in each task, as they are a reflection of a realistic training and evaluation scenario. Since all methods are evaluated under the same conditions, the comparison is equivalent. Lastly, the Stream dataset contains a diversity of images, reflected by the curation protocol used to compose each task dataset.

\subsection{Training Configurations}
\label{sec:exp_details}

We report the performance of SGD on each task in \cref{table:dataset} and \cref{table:dataset_2}. SGD performance can be considered as an upper bound for our model and train configuration. We compute SGD performance by training on the task in an isolated manner and without applying any method to mitigate forgetting. We evaluate SGD performance on the validation set of each task. 

When evaluating the time performance of each method we compute the total time for the method to train on a task. The total duration of the training run can include a warm-up and post-train phase as part of each method. We use CIFAR-100 as an auxiliary dataset for DMC\cite{DMC}.

The train configuration is reported in \cref{table:train_param}. We reset the learning rate before and after each rehearsal episode. For both the baseline methods and our method, we use an MLP with residual connections as a backbone model. We apply Batch Normalization \cite{batch_norm} after every layer and the ReLU activation function \cite{relu}. All experiments run on identical hardware of a V100 GPU cluster and we distribute the workload using Ray \cite{ray}. 

We provide in \cref{table:hparam_baseline} and \cref{table:hparam_bmc} the hyper-parameters used in Stream Benchmark experiments. We use the reported hyper-parameters for all baseline methods. When experimenting BMC on split CIFAR-100 and split Tiny-ImageNet, we use a memory size of 2,000 and a buffer size of 200 with all other settings the same as learning on Stream Benchmark.

\begin{table}[ht]
\centering
\begin{tabular}{|c|c|}
 \hline
 \hline
 \bf Benchmark & Num. tasks \\
 \hline
 Optimizer & SGD \cite{SGD}   \\
 Rehearsal Scheduler & ReduceLROnPlateau\footnotemark{}  \\
 Learning Rate & 0.1 \\
 Train epochs per Task & 2 \\
 Rehearsal epochs per Task & 100 \\
 GPU & V100 \\
 Model & Residual MLP \\
 Res. Blocks & 2 \\
 Res. Layers & 3 \\
 Dropout & 0.3 \\
 Res. Dim & 256 \\
 Hidden Dim & 128 \\
 Initialization & Xavier \cite{xavier} \\
 \hline
 \hline
\end{tabular}
\caption{Train Configuration. Detailed overview in \cref{sec:exp_details}.}
\label{table:train_param}
\end{table}
\footnotetext{\url{https://pytorch.org/docs/stable/generated/torch.optim.lr_scheduler.ReduceLROnPlateau.htm}}

\begin{table}[h]
\centering
\begin{tabular}{ | l | l | r |}
\hline
\hline
  \textbf{Baselines} & Hyper-params & Values  \\ 
  \hline
 & Memory size & 10,000 \\
 \hline 
 ER & Replay coef. & 1.0 \\
 \hline 
 DER & Distill coef. & 0.5 \\
 \hline
 \multirow{2}{*}{DER++} & Replay coef. & 1.0 \\
 & Distill coef. & 0.5 \\
 \hline
 \multirow{3}{*}{GDumb} & Max/min LR & 5e-2 / 5e-4 \\
 & Epochs & 256 \\
 & Cutmix & None \\
 \hline
 \multirow{3}{*}{HAL} & Penalty coef. & 0.1 \\
 & Beta & 0.5 \\
 & Gamma & 0.1 \\
 \hline
 \multirow{2}{*}{LwF} & Penalty coef. & 0.5 \\
 & Temperature & 2.0 \\
 \hline
 \multirow{2}{*}{GSS} & Minibatch size & 10 \\
 & Batch Num & 1 \\
 \hline
 \multirow{2}{*}{DMC} & Consolidation LR & 0.05 \\
 & Consolidation Epochs & 10 \\
 \hline
 \multirow{2}{*}{EWC$_\text{on}$} & Penalty coef. & 0.7 \\
 & Gamma & 1.0 \\
 \hline
 \multirow{2}{*}{MAS} & Penalty coef. & 0.7 \\
 & Gamma & 1.0 \\
 \hline
 \multirow{2}{*}{SI} & c & 0.5 \\
 & xi & 1.0 \\
\hline
\hline
\end{tabular}
\caption{Hyper-parameter settings for training the baseline methods on Stream Benchmark. Baselines that do not have extra hyper-parameters are omitted (A-GEM, iCaRL).}
\label{table:hparam_baseline}
\end{table}

\begin{table}[h]
\centering
\begin{tabular}{ | l | l | r |}
\hline
\hline
 \textbf{Phase} & Hyper-params & Values  \\ 
 \hline
 \multirow{2}{*}{Regularization} & Stability coef. & 1.0 \\
 & Buffer size & 10,000 \\
 \hline
 \multirow{4}{*}{Consolidation} & Num. Experts & 10 \\
 & Task loss coef. & 1.0 \\
 & Consolidation coef. & 1.0 \\
 & Buffer sampling & Random \\
\hline
\hline
\end{tabular}
\caption{Hyper-parameter settings for training BMC on Stream Benchmark.}
\label{table:hparam_bmc}
\end{table}

\section{Stream Benchmark Analysis}
\label{sec:benchmark}
\begin{table}[H]
\centering
\begin{tabular}{|c| c| c|}
 \hline
 \hline
 \bf Methods & Stream Benchmark $\uparrow$ & Time $\downarrow$
 \tabularnewline 
 \hline
 SGD & 2.1  & 100\%
 \tabularnewline 
 Multi-Task &   89.3 & 100\%
 \tabularnewline
 \hline
 
 AGEM \cite{AGEM}  & 6.6 & 231\% \\
AGEM$_R$ \cite{agem-r} &      4.3 &       230\%  \\
DER \cite{DER}   &      5.6 &       170\% \\
 DER++ \cite{DER}  & 19.4 &  205\%\\
 DMC \cite{DMC} & 1.0 & 140\% \\
 ER \cite{ER} & 41.4 &  184\% \\ 
ER-ACE \cite{ER} &     29.3 &       184\%  \\
 $\text{EWC}_{\text{on}}$ \cite{oEWC}  & 2.1 & 172\% \\ 
 GDumb \cite{gdumb} & 33.0 & 129\% \\
GSS  \cite{GSS}  &     - &      1203\%\footnotemark[2]  \\
HAL \cite{HAL}   &     - &       730\%\footnotemark[2]  \\
 iCaRL \cite{iCaRL}  & 23.4 & 141\% \\ %
LwF \cite{LWF}   &     - &       201\%\footnotemark[2]  \\
MAS \cite{MAS} &  2.1 & 144\% \\
 SI \cite{SI} & 1.4 & 515\% \\ 
\hline
 \textbf{BMC} ({Ours}) & \textbf{70.4} &  \textbf{78\%}  \\

 \hline
 \hline
\end{tabular}
\caption[Caption for Results Table]{CIL performance on Stream Dataset for 71 tasks. We use `-' to denote results that were not feasible to obtain for all 71 tasks due to intractable runtime. \footnotemark[2]Signifies incomplete time-performance evaluation. Detailed explanation of the results in \cref{sec:benchmark}}
\label{table:benchmark_sup}
\end{table}

We evaluate a set of recent (\eg DER++ \cite{DER}) and old baselines (\eg ER \cite{ER}) applicable to our setting, while some other recent baselines have a limited setting, i.e., Transformer models \cite{l2p}. Recent methods achieved better performance on standard benchmarks (i.e. CIFAR-100) did not outperform a naive baseline (ER) on Stream.

Some methods such as GSS \cite{GSS} can have an intractable run-time that grows with the number of tasks learned. Other methods, such as LwF \cite{LWF}, have a warm-up stage that requires using the train dataset. Such methods fail to complete past Task id 34 (iNaturalist \cite{DS:inaturalist}) since the step of the method is coupled with the size of the task. iNaturalist \cite{DS:inaturalist} is made up of 2686843 images, so if a step of a method requires constructing a buffer with new artifacts for each sample \cite{LWF} or performing multiple back-props \cite{GSS}, the method may not complete the task. The time-performance factor in \cref{table:benchmark_sup} may not reflect the failure, such as for LwF \cite{LWF}, since the factor is calculated at the end of training on a task.

We allow all baselines to run uninterrupted for 4 days and we terminate the experiment on the last day. All methods for which we report results finish the benchmark within 24 hours. 

\subsection{Normalized Performance}
\begin{figure*}[ht]
  \centering
  \begin{subfigure}{0.5\linewidth}
  \includegraphics[width=1.0\linewidth]{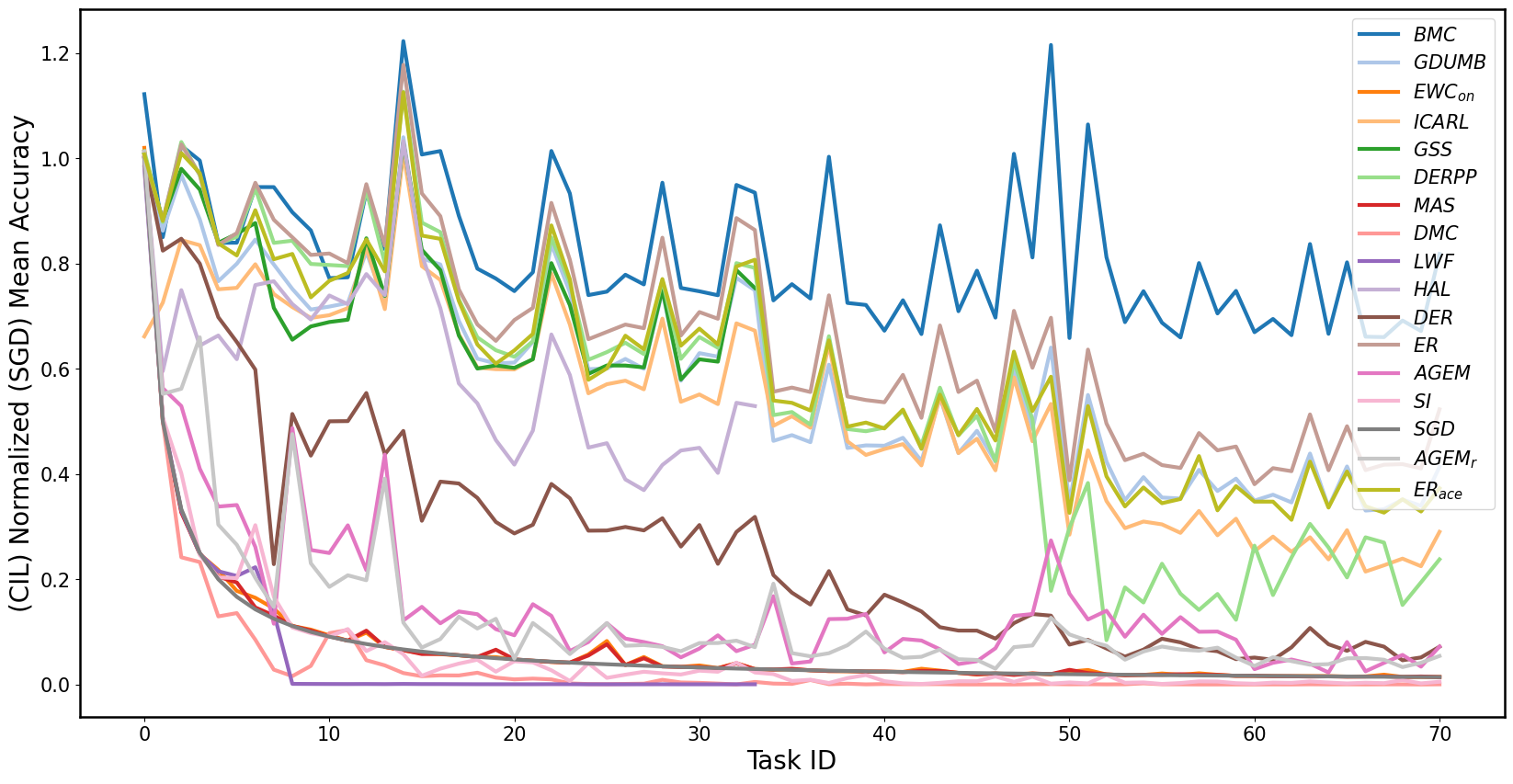}
\end{subfigure}
  \begin{subfigure}{0.49\linewidth}
  \includegraphics[width=1.0\linewidth]{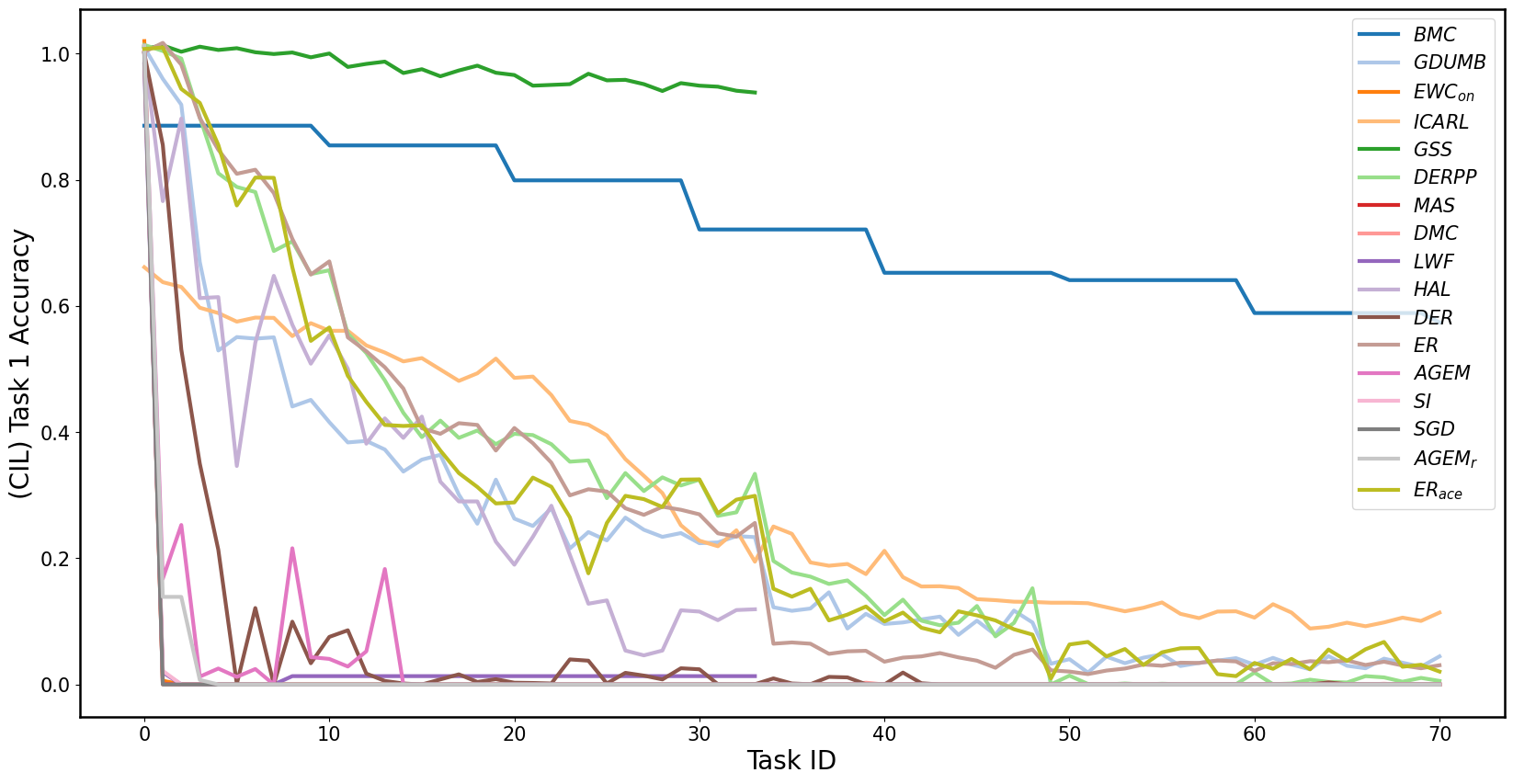}
  \end{subfigure}
\caption{\textbf{Left} mean accuracy normalized for task difficulty. Easier tasks are less difficult to forget. \textbf{Right} performance on the first task. GSS maintains performance on the first task but fails to learn new tasks \cref{fig:mean_acc}.}
\label{fig:mean_acc_aux}
\end{figure*}

\cref{fig:mean_acc_aux} presents the normalized performance by task difficulty as discussed in \cref{sec:dataset}. It can be observed that there is noise when normalizing for task difficulty, or that the method performs better than the upper bound for that task. The results can be explained as an artifact of forward transfer \cite{Survey:DefyingForgetting} as well as inherited problems with the task dataset, such as overfitting or an imbalanced train-validation split \cite{DS:dtd}. 

\subsection{First Task Performance}
\begin{figure*}
  \centering
  \includegraphics[width=1.0\linewidth]{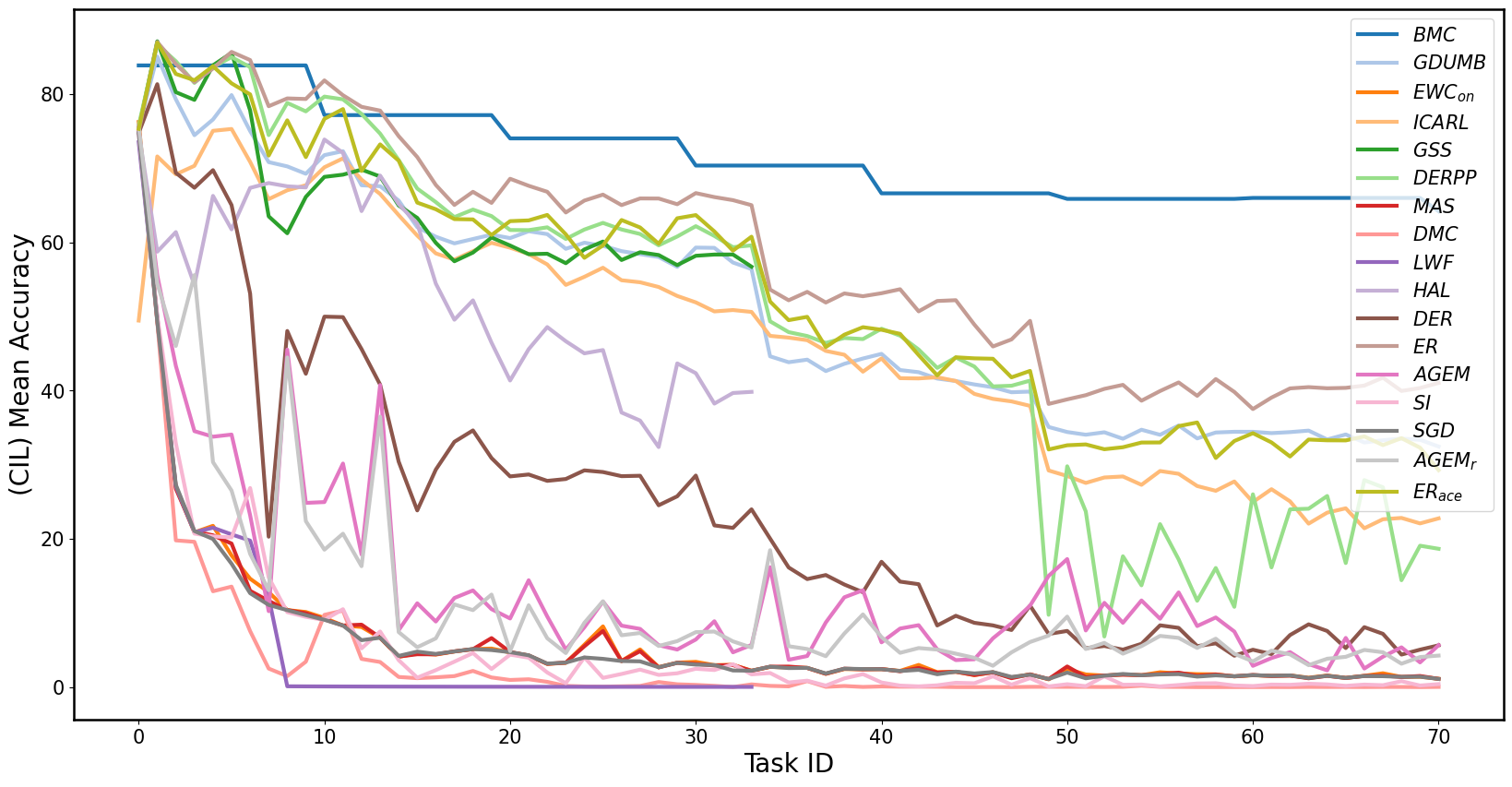}
\caption{(CIL) Mean Accuracy, some methods (GSS, HAL, LwF) \cite{GSS,HAL,LWF} fail to complete task 34 (iNaturalist \cite{DS:inaturalist}) due to the size of the dataset.}
\label{fig:mean_acc}
\end{figure*}

When evaluating the performance of a method on first task, we observe similar results to the mean task accuracy (\cref{fig:mean_acc}), with some exceptions. BMC, our method retains performance on the first task as compared to other baselines. Additionally, the performance difference between mean accuracy and first task accuracy between our method and the baselines is more prominent when evaluating only on the first task. GSS \cite{GSS}, retains performance on the first task but does not learn new tasks. iCaRL \cite{iCaRL} outperforms other baselines on the first task accuracy, but it does not perform as well when evaluated in mean accuracy. We hypothesize that this is due to the herding strategy used by iCaRL to compose the buffer that can avoid the task-recency bias.

\subsection{Time Performance Evaluation}
\begin{figure*}
  \centering
  \includegraphics[width=1.0\linewidth]{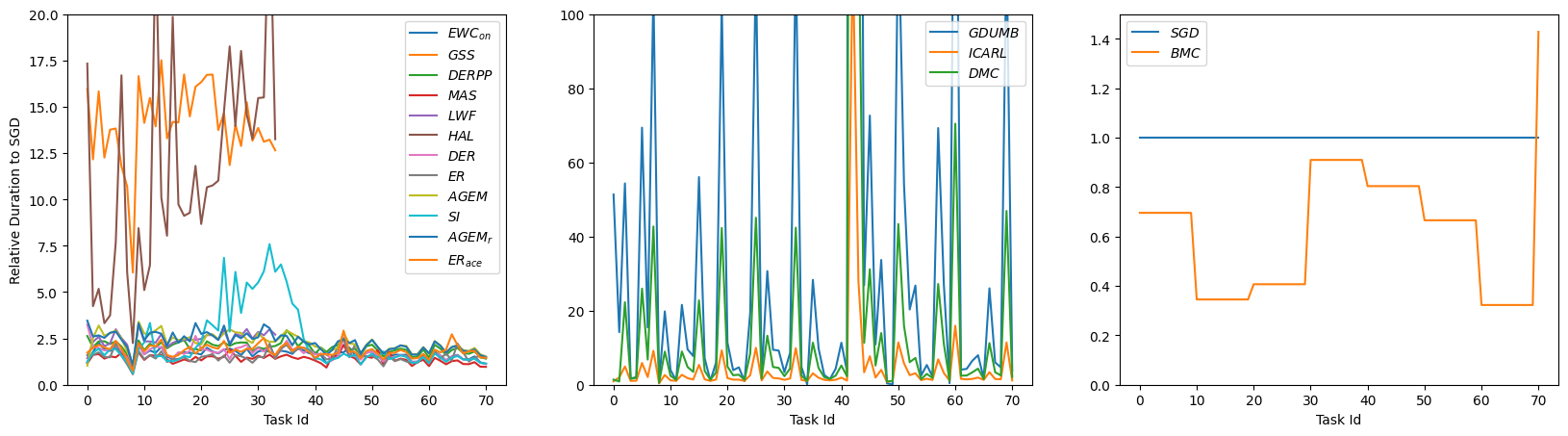}
\caption{Relative time performance of each baseline compared to run-time using SGD (\textbf{Left}). A factor of 1 signifies equivalent performance to SGD. SGD performance is independent of the task dataset size, while some methods can have a component that depends on the size of the task dataset (\textbf{Middle}). Our method (\textbf{Right}) performs better than SGD under a distributed setting. }
\label{fig:relative_perf_per_task}
\end{figure*}

Most methods perform similarly in terms of run-time on the benchmark. Some notable exceptions are SI \cite{SI}, GSS \cite{GSS} and HAL \cite{HAL}. Other methods can have a run-time performance that can be seen as non-equivalent. Such methods can perform a step that is agnostic to the task dataset size, such as constructing an auxiliary dataset \cite{DMC} or training on an auxiliary buffer \cite{gdumb,iCaRL}. As such, the run-time performance of such methods can fluctuate greatly between tasks \cref{fig:relative_perf_per_task} (Middle). Additionally, the relative time performance \cref{table:benchmark_sup} can appear inflated.

Our method, BMC, can take advantage of training multiple tasks in a distributed fashion and, as such, perform better than the baseline SGD \cref{fig:relative_perf_per_task} (\textbf{Middle}). Our method has a run-time bottleneck by the largest task in each train incremental step. This is due to the wait operation between each task-batch in order to apply batched distillation loss. For the last task of the 71 datasets, our method is slower than SGD as the performance effect of batched-task incremental learning is not utilized.

\subsection{Pareto Front}
\label{sec:pareto}
The Pareto front of our method shows a trade-off between the Total Cost of a memory and a buffer with Mean Accuracy. We run multiple experiments and vary the buffer size and memory size. We independently sample the Buffer and Memory size configuration between 8k to 20k exemplars at the start of each experiment. The Total Cost represents the mean number of exemplars stored at each train incremental step of our method. The performance gain plateaus as we increase $\mathbb{TC}$. Some configurations are not Pareto optimal, such as the use of a very small buffer and really large memory or vice versa. Both the buffer and memory are integral parts of our method's performance but observe a limitation on the improvement they provide to the performance beyond a certain point. As such we hypothesize that improvements in both the utilization and construction of the buffer and memory are more significant than the size of the buffer. 

Lastly, we motivate a method not to be evaluated at a single point when evaluating the cost performance of the method. As can be observed, the relationship between $\mathbb{TC}$ and Mean Accuracy is not linear and requires that more than one configuration be evaluated.

\begin{figure}[H]
  \centering
  \includegraphics[width=1.0\linewidth]{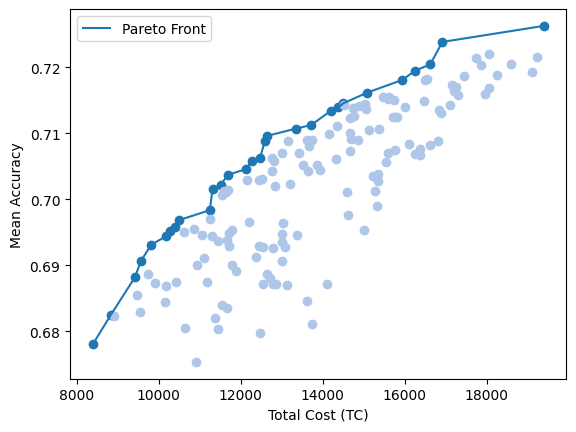}
\caption{Trade-off between $\mathbb{TC}$ and Mean Accuracy for our method (BMC) with details discussed in \cref{sec:pareto}. }
\label{fig:pareto}
\end{figure}

\section{Ablation Experiments}
\label{sec:ablation}
\begin{figure*}[ht]
  \centering
  \begin{subfigure}{1\linewidth}
  \includegraphics[width=1.0\linewidth]{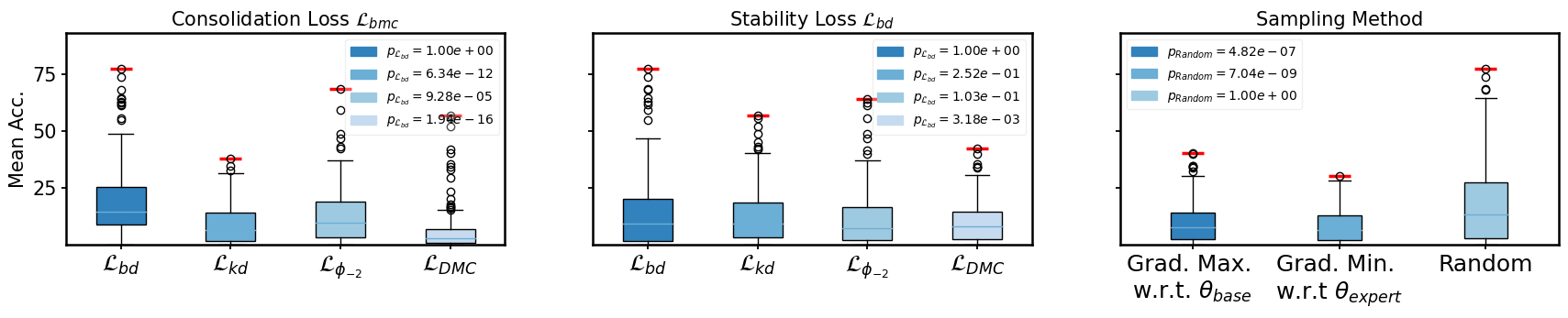}  
  \end{subfigure}
  \begin{subfigure}{1\linewidth}
  \includegraphics[width=1.0\linewidth]{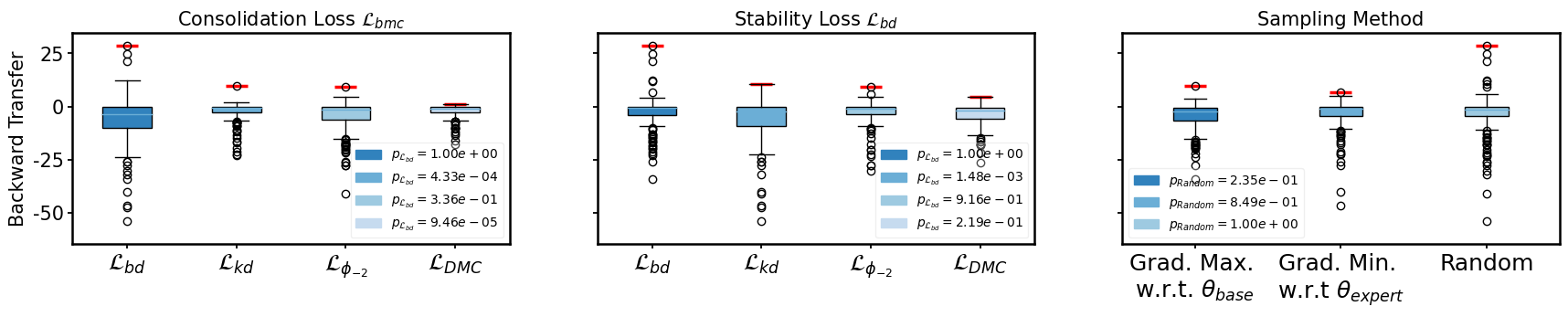}
\end{subfigure}
  
\caption{
Aggregated results on an ablation study with 629 experiments for Permuted-MNIST. We randomly sample the type of loss for each phase of our method, the weight coefficient of each component as well as the sampling method. From Left to Right. Evaluation of loss in direct replacement to $\mathcal{L}_{bd}$ (Ours) for alternative losses $\mathcal{L}_{kd}$, $\mathcal{L}_{\phi_{-2}}$, $\mathcal{L}_{DMC}$. Stability Loss as a regularization component in replacement to  $\mathcal{L}_{bd}$ and sampling method in replacement to Random. (\textbf{Top}) row reports the Mean Accuracy, where higher is better. (\textbf{Bottom}) row reports the Backward Transfer, where higher is better. $\mathcal{L}_{bd}$ has both the best-performing trials and a higher mean score for each metric. Naive random sampling for constructing the buffer performs the best. $p$ values are reported from one-way ANOVA between our purposed method and each corresponding method. Additional details in \cref{sec:lbmc}, \cref{sec:lbd} and \cref{sec:sampling}}

\label{fig:ablation_sup}
\end{figure*}

We motivate our ablation study in performing an unbiased estimate of each component of our method. Using the benchmark dataset can be a biased estimate of a component of a method for the following reasons. First, we need a dataset for which we have access to a large number of tasks that are evenly divisible by the number of experts we evaluate, i.e. 128 by 16. Second, for the benchmark the tasks have different domain-gaps, for example the difference between `Lego' and `Rooms' datasets. Third, the tasks have varying lengths and numbers of classes, such that `Lego' has 32,000 train images with 46 classes and `Rooms' has 3,937 images with 5 classes. An ablation study on a dataset with multiple sources of experimental variance requires additional experimental trials for an unbiased estimate. Permuted-MNIST meets all of the above requirements but for the same reasons is not suitable for a benchmark. 

We run 629 experiments of 128 tasks and randomly sample each hyper-parameter that controls a different component for our method, which require 1 week of training time on a GPU cluster of x8 V100. We vary the stability coefficient ($\lambda$), consolidation coefficient ($\beta$), Number of Experts, Consolidation Loss $\mathcal{L}_{bmc}$, Stability Loss $\mathcal{L}_{bd}$ and both Memory and Buffer sampling Method. Each experimental configuration is randomly sampled, and as such, it is important to consider both the mean and the best-performing configuration when evaluating a setting. The reason is that there can be poor synergy between two randomly sampled settings or that a method is not fully evaluated. For example, consider that a really small value for $\lambda$ can be used and as such the full effect of the loss function used for a stability loss cannot be evaluated in that context. Additionally, each Stability Loss can have different sensitivity to $\lambda$ and as such maintaining the coefficient fixed or evaluating on different ranges can make the comparison non-equivalent. All settings have their values randomly sampled from the reported interval. Results presented in \cref{fig:ablation_sup}, \cref{fig:ewc}, \cref{fig:param} and discussed in this section.

\subsection{Batched Distillation Loss}
\label{sec:lbmc}
We consider several alternatives in direct replacement to $\mathcal{L}_{bd}$. We use $\mathcal{L}_{bd}$ for two components of our method, on the Consolidation Loss for $\mathcal{L}_{bmc}$ and as a Stability Loss. We report the results in \cref{fig:ablation_sup} (Left and Middle). We evaluate three alternative methods, such as $\mathcal{L}_{kd}$, $\mathcal{L}_{\phi_{-2}}$, $\mathcal{L}_{DMC}$.
$\mathcal{L}_{kd}$ is Knowledge Distillation (KD) \cite{KnowledgeDistillation} applied on the logit space, similar to \cite{DER, iCaRL}. 
$\mathcal{L}_{\phi_{-2}}$ is Knowledge Distillation applied on the pen-ultimate representation \cite{KD-penult, KD-feature}. $\mathcal{L}_{DMC}$ is Knowledge Distillation applied on a slice of logits using \textit{double distillation loss} \cite{DMC}, similar to \cite{ExModel}. We evaluate the statistical significance on both evaluation metrics, Mean Accuracy and Backward Transfer. When considering the statistical significance on both metrics, $\mathcal{L}_{bd}$ outperforms other alternatives when used in $\mathcal{L}_{bmc}$ and as a Stability Loss. $\mathcal{L}_{\phi_{-2}}$, performs similarly to $\mathcal{L}_{bd}$ and is able to reach a higher Mean Accuracy in the study. However, we find that the results are not consistent and the mean value for $\mathcal{L}_{\phi_{-2}}$ on each metric is lower. The result is not statistically significant based on a $p$-value $>$ 0.05. As such, the two methods can be evaluated further in future work. 

\subsection{Regularization Loss}
\label{sec:lbd} 
\begin{figure}[H]
  \centering
\begin{subfigure}[h]{.5\linewidth}
  \includegraphics[width=\linewidth]{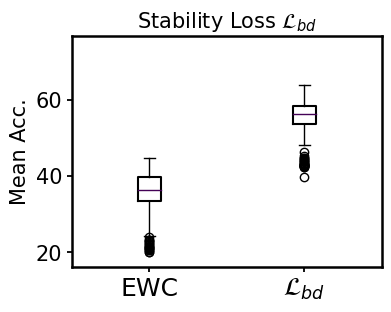}
\caption{Mean Accuracy}
\end{subfigure}
\begin{subfigure}[h]{.49\linewidth}
  \includegraphics[width=\linewidth]{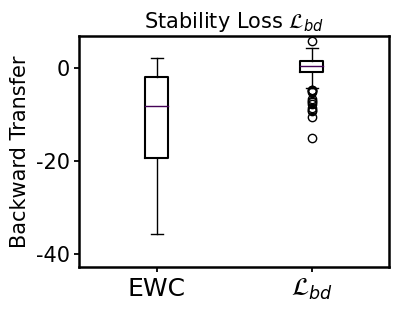}
\caption{Mean Forgetting}
\end{subfigure}
\caption{Comparison of using EWC in replacement to $\mathcal{L}_{bd}$ for Mean Accuracy (\textbf{Left}) and Backward Transfer (\textbf{Right}). }
\label{fig:ewc}
\end{figure}

Elastic Weight Consolidation (EWC) \cite{EWC} uses an alternative loss term to the current task loss that provides an optimization constraint on the parameters when training on a new task. The importance of each parameter to the current task is calculated based on an approximation of the Fisher Information Matrix. We use EWC as \textit{stability-loss} in direct replacement for $\mathcal{L}_{bd}$. \Cref{fig:ewc} shows that EWC poses a strict constraint to the parameter and is unable to learn new tasks. $\mathcal{L}_{bd}$ outperforms EWC in this context. This could be explained by the large domain shift between each permutation and the limitation in the capacity of the backbone model for which it is not possible to isolate all parameters while learning new tasks. As such, we hypothesize that constructive interference methods such as KD are better candidates for both components of our method.

\subsection{Buffer Sampling}
\label{sec:sampling}
We examine two alternatives to Random sampling. We use gradient information as a heuristic when constructing the Buffer. In detail, we use the samples that produce the largest gradient with respect to the base model ($\theta_{base}$) with the intuition that they will be the most informative during the consolidation of $\theta_{base}$. We also use samples that produce the smallest gradient norms \wrt the expert model ($\theta_{expert}$) with the intuition that they are the most representative of the task the expert was trained on. Both methods perform poorly as compared to Random sampling. We hypothesize alternatives or improvements in the buffer sampling method can outperform Random in terms of task performance, but also consider that they can perform poorly in terms of run-time.

\subsection{Parameter Importance}
\begin{figure}[H]
  \centering
  \includegraphics[width=1.0\linewidth]{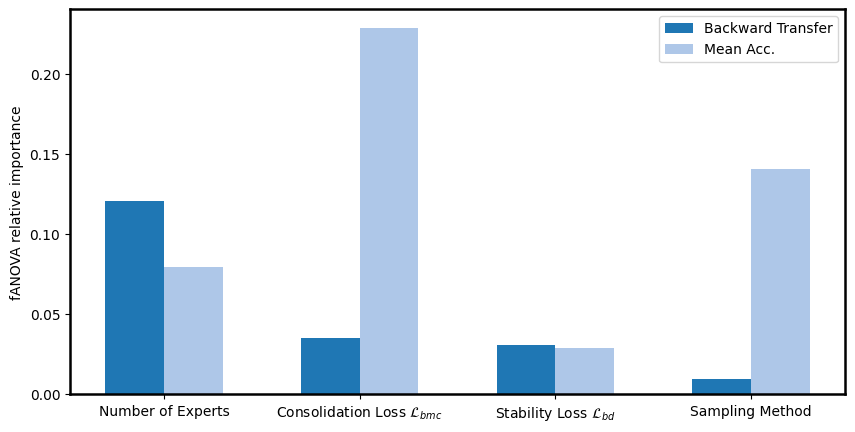}
\caption{Importance of each component of our method in Backward Transfer (Dark Blue) and Mean Accuracy (Light Blue). We report the importance score using fANOVA \cite{fanova}}
\label{fig:param}
\end{figure}

We evaluate the most important component of our method using fANOVA parameter importance \cite{fanova}. We find that the Number of Experts contributes the most to both the Mean Accuracy and Backward Transfer. Interestingly, both $\mathcal{L}_{bmc}$ and the sampling method contribute more to the Mean Accuracy than Backward Transfer. Our findings in \cref{fig:param} agree with our analysis that $\mathcal{L}_{bmc}$ provides a better approximation to the multi-task gradient as opposed to single-task consolidation and finally achieves higher Mean Accuracy within a batch of tasks. Likewise, a higher Number of Experts puts more constraints on the gradient updates than a small Number of Experts, making the gradient less `sharp'. It reduces the bias toward every single task and benefits the Backward Transfer as it protects the parameters for previous tasks.

\begin{figure}[ht]
  \centering
  \includegraphics[width=1.0\linewidth]{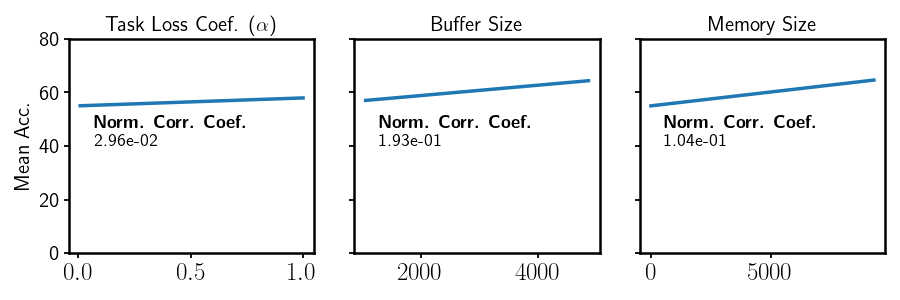}
  \caption{We vary the coefficient of task loss as well as the buffer and memory size to examine their effects on final average accuracy performance.}
  \label{fig:rebuttal_ablation}
\end{figure}

\subsection{Task Loss Coefficient \& Buffer-Memory Sizes}
We attach the ablation study results for task loss coefficient in the consolidation phase, buffer size and memory size in \cref{fig:rebuttal_ablation}. Figures for buffer and memory size are as supplementary to the correlation we reported in the main text. In task loss coefficient, we find its low correlation to the final average accuracy.

\subsection{Backbone Model}
\begin{table}[h]
\centering
\begin{tabular}{ | c | c c c c |}
\hline
\hline
  Method & CLIP & ViT & ResNet50 & Avg. \\ 
  \hline
 ER & 41.4 & 32.8 & 27.6 & 33.9 \\  
 DER++ & 19.4 & 15.1 & 12.7 & 15.7 \\
 \textbf{BMC} & \textbf{70.4} & \textbf{60.2} & \textbf{47.0} & \textbf{59.2} \\
\hline
\hline
\end{tabular}
\caption{Comparing BMC (ours), ER and DER++ on different backbone models CLIP, ViT \cite{vit} and ResNet50 on Stream Benchmark following the same experiment configurations. BMC outperforms other methods constantly on all backbones.}
\label{table:backbone}
\end{table}

Results obtained in \cref{table:benchmark_sup} are subject to the backbone model used in extracting feature vectors. As we have to optimize 16 baselines; for 6,770,722 images and 2,866 classes, CLIP embeddings allow us to evaluate the merits of each baseline without extra computational cost. We include the results of the ablation between the pre-trained CLIP, ViT and ResNet50 features on Stream Benchmark and show that our method can work across different backbones in \cref{table:backbone}. BMC outperforms the next best method (ER) by 25.3\% on 3 backbones in average. We emphasize that the evaluation of the backbone model is orthogonal to both our method and the benchmark, as any backbone can be used in direct replacement.

\clearpage
\begin{table*}[h]

\centering
\begin{tabular}{llrrrlr}
\toprule
                    Task ID &            Name &  Num. Classes  &  Num. Train Images &  Num. Val Images &         Sub Task  & Val. Acc. (SGD) \\
\midrule
\toprule
0  &            Aircraft\cite{DS:aircraft} &           70 &        3334 &      3333 &            Family &  74.74 \\
1  &              Apparel\cite{DS:apparel} &            6 &        8538 &      2847 &             Color &  98.63 \\
2  &          Aptos2019\cite{DS:aptos2019} &            5 &        2746 &       916 &                 - &  81.88 \\
3  &                      Art\cite{DS:art} &           14 &       72009 &     24004 &                 - &  84.21 \\
4  &                      Asl\cite{DS:asl} &           29 &       65250 &     21750 &                 - &  99.89 \\
5  &                    Boat\cite{DS:boat} &            9 &        2193 &       731 &                 - &  99.86 \\
6  &                    Cars\cite{DS:cars} &          196 &        8144 &      8041 &                 - &  88.70 \\
7  &            Cataract\cite{DS:cataract} &            4 &         901 &       301 &                 - &  88.70 \\
8  &                CelebA\cite{DS:celeba} &            2 &      151949 &     50650 &            Shadow &  93.42 \\
9  &        Colorectal\cite{DS:colorectal} &            8 &        7500 &      2500 &                 - &  97.16 \\
10 &            Concrete\cite{DS:concrete} &            2 &       30000 &     10000 &                 - &  99.90 \\
11 &                Core50\cite{DS:core50} &           50 &      123649 &     41217 &            Object &  99.69 \\
12 &                      Cub\cite{DS:cub200} &          200 &        5994 &      5794 &                 - &  82.31 \\
13 &        Deepweedsx\cite{DS:deepweedsx} &            9 &       15007 &      2501 &                 - &  93.24 \\
14 &              Dermnet\cite{DS:dermnet} &           23 &       15557 &      4002 &                 - &  63.09 \\
15 &                      Dtd\cite{DS:dtd} &           47 &        1880 &      1880 &           Split 1 &  76.60 \\
16 &        Electronic\cite{DS:electronic} &           36 &       16152 &      5384 &                 - &  76.10 \\
17 &                Emnist\cite{DS:emnist} &           47 &      112800 &     18800 &          Balanced &  86.61 \\
18 &              Eurosat\cite{DS:eurosat} &           10 &       20250 &      6750 &                 - &  97.61 \\
19 &                  Event\cite{DS:event} &            8 &        1180 &       394 &                 - & 100.00 \\
20 &                    Face\cite{DS:face} &            3 &       10902 &      3634 &                 - &  98.98 \\
21 &              Fashion\cite{DS:fashion} &            5 &       33329 &     11112 &            Gender &  94.47 \\
22 &              Fer2013\cite{DS:fer2013} &            7 &       28709 &      7178 &                 - &  72.99 \\
23 &                  Fgvc6\cite{DS:fgvc6} &          251 &      118475 &     11994 &                 - &  79.31 \\
24 &                    Fish\cite{DS:fish} &            9 &        6750 &      2250 &                 - & 100.00 \\
25 &              Flowers\cite{DS:flowers} &           17 &        1020 &       340 &           Split 1 &  99.12 \\
26 &              Food101\cite{DS:food101} &          101 &       75750 &     25250 &                 - &  95.03 \\
27 &            Freiburg\cite{DS:freiburg} &           25 &        3710 &      1237 &                 - &  97.33 \\
28 &            Galaxy10\cite{DS:galaxy10} &           10 &       13302 &      4434 &                 - &  77.60 \\
29 &              Garbage\cite{DS:garbage} &           12 &       11636 &      3879 &                 - &  98.20 \\

\bottomrule
\end{tabular}

\caption{Dataset used in Stream benchmark. Where applicable, Sub Task refers to the dataset split used in classifying each dataset. Additional details in \cref{sec:dataset} and additional dataset in \cref{table:dataset_2}. }
\label{table:dataset}
\end{table*}
\begin{table*}[h]

\centering
\begin{tabular}{llrrrlr}
\toprule
                    Task ID &            Name &  Num. Classes  &  Num. Train Images &  Num. Val Images &         Sub Task  & Val. Acc. (SGD) \\
\midrule
\toprule
30 &                  Gtsrb\cite{DS:gtsrb} &           43 &       39209 &     12630 &                 - &  94.13 \\
31 &            Ham10000\cite{DS:ham10000} &            7 &       15022 &      5008 &                 - &  95.09 \\
32 &      Handwritten\cite{DS:handwritten} &           33 &        1237 &       413 &           Letters &  74.09 \\
33 &        Histaerial\cite{DS:histaerial} &            7 &       26460 &     11340 &             Small &  75.26 \\
34 &      iNaturalist\cite{DS:inaturalist} &           51 &     2686843 &    100000 &           Species &  96.36 \\
35 &                Indoor\cite{DS:indoor} &           67 &        5360 &      1340 &                 - &  92.46 \\
36 &                  Intel\cite{DS:intel} &            6 &       14034 &      3000 &                 - &  95.90 \\
37 &                    Ip02\cite{DS:ip02} &          102 &       52603 &     22619 &                 - &  70.14 \\
38 &      Kermany2018\cite{DS:kermany2018} &            4 &       83516 &       968 &                 - &  97.00 \\
39 &  Kvasircapsule\cite{DS:kvasircapsule} &           14 &       28342 &      9448 &                 - &  97.52 \\
40 &              Landuse\cite{DS:landuse} &           21 &        9450 &      1050 &                 - &  99.05 \\
41 &                    Lego\cite{DS:lego} &           46 &       32000 &      8000 &                 - &  91.20 \\
42 &              Malacca\cite{DS:malacca} &            3 &         121 &        41 &                 - & 100.00 \\
43 &                  Manga\cite{DS:manga} &            7 &         341 &       114 &                 - &  76.32 \\
44 &            Minerals\cite{DS:minerals} &            7 &        4111 &      1371 &                 - &  93.87 \\
45 &                Office\cite{DS:office} &           65 &        1820 &       607 &               Art &  84.68 \\
46 &                Oriset\cite{DS:oriset} &            4 &       11110 &      3703 &           Origami &  95.52 \\
47 &                Oxford\cite{DS:oxford} &           17 &        3797 &      1266 &                 - &  66.03 \\
48 &                    Pcam\cite{DS:pcam} &            2 &      262144 &     32768 &                 - &  82.06 \\
49 &          Places365\cite{DS:places365} &          365 &     1803460 &     36500 &                 - &  54.79 \\
50 &              Planets\cite{DS:planets} &           11 &        1228 &       410 &                 - & 100.00 \\
51 &            Plantdoc\cite{DS:plantdoc} &           28 &        2340 &       236 &                 - &  61.86 \\
52 &          Pneumonia\cite{DS:pneumonia} &            2 &        5232 &       624 &                 - &  81.09 \\
53 &              Pokemon\cite{DS:pokemon} &          150 &        4994 &      1665 &                 - &  95.62 \\
54 &            Products\cite{DS:products} &           12 &       59551 &     60502 &                 - &  88.10 \\
55 &            Resisc45\cite{DS:resisc45} &           45 &       23625 &      7875 &                 - &  95.77 \\
56 &                    Rice\cite{DS:rice} &            5 &       56250 &     18750 &                 - &  99.84 \\
57 &                    Rock\cite{DS:rock} &            7 &        1515 &       506 &                 - &  82.21 \\
58 &                  Rooms\cite{DS:rooms} &            5 &        3937 &      1313 &                 - &  93.37 \\
59 &                      Rvl\cite{DS:rvl} &           16 &      320000 &     39999 &                 - &  88.03 \\
60 &                  Santa\cite{DS:santa} &            2 &         614 &       616 &                 - &  98.54 \\
61 &          Satellite\cite{DS:satellite} &            4 &       35679 &     11894 &                 - &  94.97 \\
62 &            Simpsons\cite{DS:simpsons} &           42 &       31399 &     10467 &                 - &  99.38 \\
63 &                Sketch\cite{DS:sketch} &          250 &       15000 &      5000 &                 - &  78.82 \\
64 &                Sports\cite{DS:sports} &          100 &       14072 &       500 &                 - &  99.00 \\
65 &                    Svhn\cite{DS:svhn} &           10 &       73257 &     26032 &                 - &  82.23 \\
66 &            Textures\cite{DS:textures} &           64 &        4335 &      4340 &                 - &  99.82 \\
67 &          Vegetable\cite{DS:vegetables} &           15 &       18000 &      3000 &                 - &  99.93 \\
68 &      Watermarked\cite{DS:watermarked} &            2 &       24987 &      6588 &                 - &  95.39 \\
69 &              Weather\cite{DS:weather} &            4 &         841 &       281 &                 - &  98.22 \\
70 &              Zalando\cite{DS:zalando} &            6 &       24270 &      8090 &                 - &  78.44 \\
\bottomrule
\end{tabular}

\caption{Dataset used in Stream benchmark. When applicable, Sub Task refers to the split of the dataset used in classifying each dataset. Additional details in \cref{sec:dataset} and additional dataset in \cref{table:dataset}. }
\label{table:dataset_2}
\end{table*}